%% file: paper.tex
\documentclass[10pt,twocolumn,letterpaper]{article}

\usepackage{iccv}
\usepackage{times}
\usepackage{epsfig}
\usepackage{amsmath}
\usepackage{amssymb}

\usepackage{subcaption}

\usepackage{graphicx}
\usepackage{color}
\usepackage{booktabs, multirow, array, tabularx}
\usepackage[table,xcdraw]{xcolor}
\usepackage{hhline}
\usepackage{caption}
\usepackage[linesnumbered,ruled,vlined,noend]{algorithm2e}
\usepackage{setspace}

\makeatletter
\newcommand{\removelatexerror}{\let\@latex@error\@gobble}
\makeatother

\usepackage{inconsolata}

\usepackage[pagebackref=true,breaklinks=true,letterpaper=true,colorlinks,bookmarks=false]{hyperref}

\iccvfinalcopy 

\def\iccvPaperID{5274}

\newcolumntype{L}[1]{>{\raggedright\arraybackslash}p{#1}}
\newcolumntype{C}[1]{>{\centering\arraybackslash}p{#1}}

\newcommand{\jitnet}{JITNet}

\usepackage[font=small]{caption}

\definecolor{redcolor}{rgb}{0.75,0.0,0.0}
\newcommand{\red}[1]{{\color{redcolor}\textbf{\emph{#1}}}}

\begin{document}

\title{Online Model Distillation for Efficient Video Inference}

\author{Ravi Teja Mullapudi\textsuperscript{1}
\quad Steven Chen\textsuperscript{2}
\quad Keyi Zhang\textsuperscript{2}
\quad Deva Ramanan\textsuperscript{1}
\quad Kayvon Fatahalian\textsuperscript{2}
\and 
\textsuperscript{1} Carnegie Mellon University 
\and
\textsuperscript{2} Stanford University
}

\maketitle
\input{abstract}
\input{intro}

\input{related}
\input{jitnet}
\input{dataset}
\input{experiments}
\input{conclusion}
\input{ack}
\input{supp_arxiv}

{\small
\bibliographystyle{ieee_fullname}
\bibliography{ref}
}
\end{document}

%% file: abstract.tex
\begin{abstract}
High-quality computer vision models typically address the problem of
understanding the general distribution of real-world images.  However, most
cameras observe only a very small fraction of this distribution. This offers
the possibility of achieving more efficient inference by specializing
compact, low-cost models to the specific distribution of frames observed by
a single camera.  In this paper, we employ the technique of model
distillation (supervising a low-cost student model using the output of a
high-cost teacher) to specialize accurate, low-cost semantic segmentation
models to a target video stream. Rather than learn a specialized student
model on offline data from the video stream, we train the student in an
online fashion on the live video, intermittently running the teacher to
provide a target for learning.  Online model distillation yields semantic
segmentation models that closely approximate their Mask R-CNN teacher with 7~to~17$\times$ lower inference runtime cost (11~to~26$\times$ in FLOPs),
even when the target video's distribution is non-stationary.
Our method requires no offline pretraining
on the target video stream, achieves higher accuracy and lower cost than
solutions based on flow or video object
segmentation, and can exhibit better temporal stability than the original teacher.  We also provide a new video dataset for evaluating the
efficiency of inference over long running video streams.
\end{abstract}

%% file: intro.tex
\section{Introduction}

\begin{figure}[t!] \centering
\includegraphics[width=0.45\textwidth]{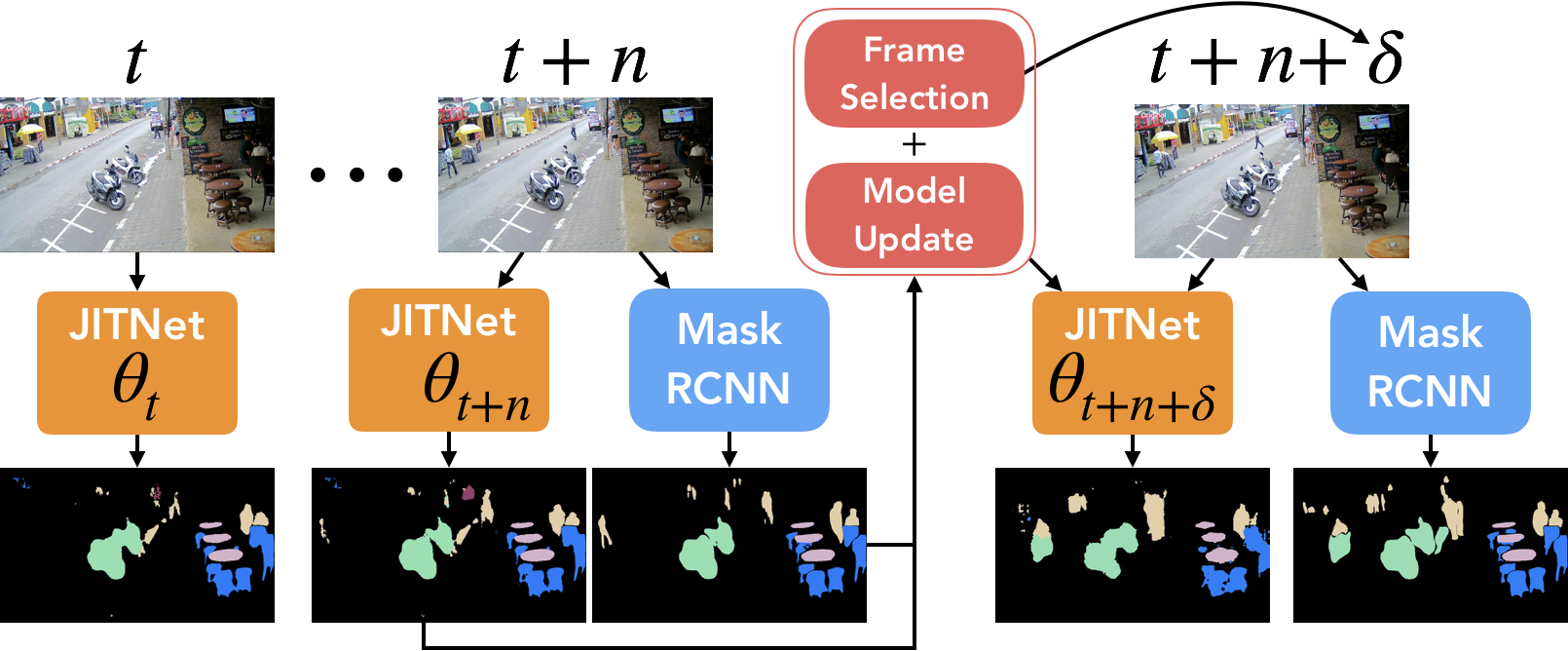}
\caption{Online model distillation overview: A low-cost student model is tasked
    to generate a high-resolution, per-frame semantic segmentation. To retain
    high accuracy, as new frames arrive, an expensive teacher model's (MRCNN)
    output is periodically used as a learning target to adapt the student and
    selecting the next frame to request supervision. We call the student
    model ``\jitnet" since is designed to be specialized ``just-in-time" for
    future frames.}
\vspace{-1.5em}
\label{fig:overview}
\end{figure}

Many computer vision algorithms focus on the problem of understanding the most
general distribution of real-world images (often modeled by ``Internet''-scale
datasets such as ImageNet\,\cite{ilsvrc15} or COCO\,\cite{lin2014microsoft}).  
In contrast, most real-world video cameras capture scenes that feature a much
narrower distribution of images, and this distribution can evolve continuously
evolve over time.  For example, stationary cameras observe scenes that evolve
with time of day, changing weather conditions, and as different subjects move
through the scene.  TV cameras pan and zoom, most smartphone videos are
hand-held, and egocentric cameras on vehicles or robots move through dynamic
scenes.

In this paper, we embrace this reality and move away from attempting to pre-train
a model on camera-specific datasets curated in advance, and instead train models
\emph{online on a live video stream as new video frames arrive}.  Specifically,
we apply this methodology to the task of realizing high-accuracy and low-cost
semantic segmentation models that continuously adapt to the contents of a video
stream.  

We employ the technique of model
distillation\,\cite{bucilua2006model,hinton2015distilling}, training a
lightweight ``student'' model to output the predictions of a larger, reliable
high-capacity ``teacher'', but do so in an online fashion, intermittently
running the teacher on a live stream to provide a target for student learning.
We find that simple models can be accurate, provided they are continuously
adapted to the specific contents of a video stream as new frames arrive (i.e.
models can learn to cheat---segmenting people sitting on a park lawn might be as
easy as looking for shades of green!).  To achieve high efficiency, we require a
new model architecture that simultaneously supports low-cost inference and fast
training, as well as judicious choice of when to periodically run the teacher to
obtain supervision.

We show that online model distillation yields semantic segmentation models that
closely approximate their Mask R-CNN\,\cite{he2017mask} teacher with 7 to
17$\times$ lower inference runtime cost (11-26$\times$ when comparing FLOPs),
even when the target video's distribution is non-stationary over time.  Our
method requires no offline pretraining on data from the target video stream, has
a small number of hyper parameters, and delivers higher accuracy segmentation
output, than low-cost video semantic segmentation solutions based on flow.  The
output of our low-cost student models can be preferable (in terms of temporal
stability) to that of the expensive teacher.  We also provide a new video
dataset designed for evaluating the efficiency of inference over long running
video streams.

%% file: related.tex
\section{Related Work}

\paragraph{Distillation for specialization:}  Training a small, efficient model
to mimic the output of a more expensive teacher has been proposed as a form of
model compression (also called knowledge distillation)\,\cite{bucilua2006model,hinton2015distilling}. While early
explorations of distillation focused on approximating the output of a large
model over the entire original data distribution, our work, like other recent
work from the systems community\,\cite{kang2017noscope}, leverages distillation
to create highly compact, domain-specialized models that need only mimic the
teacher for a desired subset of the data. Prior specialization approaches rely on
tedious configuration of
models\,\cite{liu2013intelligent,fleuret2008multicamera} or careful selection of
model training samples so as not to miss rare events\,\cite{lu2013learning}.
Rather than treating model distillation as an offline training preprocess for a
stationary target distribution (and incurring the high up-front training cost
and the challenges of curating a representative training set for each unique
video stream), we perform distillation online to adapt the student model
dynamically to the changing contents of a video stream.



\vspace{-1em}
\paragraph{Online training:}  Training a model online as new video frames arrive
violates the independent and identically distributed  (i.i.d) assumptions of
traditional stochastic gradient descent optimization.  Although online learning
from non-i.i.d data streams has been
explored~\cite{cesa2006prediction,shalev2012online}, in general there has been
relatively little work on online optimization of ``deep'' non-convex predictors
on correlated streaming data.  The major exception is the body of work on deep
reinforcement learning~\cite{mnih2015human}, where the focus is on learning
policies from experience. Online distillation can be formulated as a
reinforcement or a meta-learning\,\cite{pmlr-v70-finn17a} problem. However,
training methods\,\cite{schulman2017proximal, mnih2016asynchronous} employed in
typical reinforcement settings are computationally expensive, require a large
amount of samples, and are largely for offline use. Our goal is to train a compact
model which mimics the teacher in a small temporal window. In this context,
we demonstrate that standard gradient decent is effective for online
training our compact architecture.

\vspace{-1em}
\paragraph{Tracking:} Traditional object tracking
methods\,\cite{kalal2012tracking,hare2016struck,henriques2015high} and more
recent methods built upon deep feature
hierarchies\,\cite{ma2015hierarchical,wang2015visual,hong2015online,nam2016learning}
can be viewed as a form of rapid online learning of appearance models from
video. Tracking parameterizes objects with bounding boxes rather than
segmentation masks and its cost scales in complexity with the number of objects
being tracked. Our approach for online distillation focuses on pixel-level
semantic segmentation and poses a different set of performance challenges. It can
be viewed as learning an appearance model for the entire scene as opposed to
individual objects.


\vspace{-1em}
\paragraph{Fast-retraining of compact models:}  A fundamental theme in our work
is that low-cost models that do not generalize widely are useful, provided they
can be quickly retrained to new distributions.  Thus, our ideas bear similarity
to recent work accelerating image classification in video via online adaptation
to category skew\,\cite{Shen:2017:adaptivedetection} and on-the-fly model
training for image super-resolution\,\cite{Shocher:2018:superres}.

\vspace{-1em}
\paragraph{Video object segmentation:}  Solutions to video object segmentation
(VOS) leverage online adaptation of high-capacity deep models to a provided
reference segmentation in order to propagate instance masks to future
frames\,\cite{perazzi2017learning,yang2018efficient,voigtlaender:2017:onAVOS,caelles2017one}.
The goal of these algorithms is to learn a high-quality, video-specific
segmentation model for use on subsequent frames of a short video clip, not to
synthesize a low-cost approximation to a pre-trained general segmentation model
like Mask R-CNN\,\cite{he2017mask} (MRCNN).  VOS solutions require seconds to minutes of
training per short video clip (longer than directly evaluating a general
segmentation model itself), precluding their use in a real-time setting. We
believe our compact segmentation architecture and online distillation method
could be used to significantly accelerate top-performing VOS solutions (see
Section~\ref{sec:evaluation}).


\vspace{-1em}
\paragraph{Temporal coherence in video:} Leveraging frame-to-frame coherence in
video streams, such as background subtraction or difference detection, is a
common way to reduce computation when processing video streams.  More advanced
methods seek to activate different network layers at different temporal
frequencies according to expected rates of
change\,\cite{koutnik2014clockwork,shelhamer2016clockwork} or use frame-to-frame
flow to warp inference results (or intermediate features) from prior frames to
subsequent frames in a video\,\cite{gadde2017semantic,zhu2017deep}.  We show
that for the task of semantic segmentation, exploiting frame-to-frame coherence
in the form of model specialization (using a compact model trained on recent
frames to perform inference on near future frames)  is both more accurate and
more efficient than flow-based methods on a wide range of videos.



%% file: jitnet.tex
\section{Just-In-Time Model Distillation}
Figure~\ref{fig:overview} provides a high-level overview of online model
distillation for high quality, low-cost video semantic segmentation.
On each video frame, a compact model is run, producing a pixel-level
segmentation. This compact student model is periodically adapted using
predictions from a high-quality teacher model (such as MRCNN\,\cite{he2017mask}). Since the student model is trained online (adapted
just-in-time for future use), we refer to it as ``\jitnet''.  To
make online distillation efficient in practice, our approach must: 1) use a student network that is fast for
inference and fast for adaptation, 2) train this student \textit{online} using imperfect
teacher output, and 3) determine when and how to ask the teacher for labels as new frames arrive.
We next address each of these challenges in turn.

\subsection{JITNet Architecture}
\label{sec:jitnetArch}
Efficient online adaptation requires a student architecture that (1) is
efficient to evaluate even when producing high resolution outputs and (2) is
amenable to fast learning. The ability to make high-resolution predictions is
necessary for handling real-world video streams with objects at varying scales.
Fast and stable adaptation is necessary for learning from the teacher in a small
number of iterations. 

\begin{figure}[t!] \centering
\includegraphics[width=\linewidth]{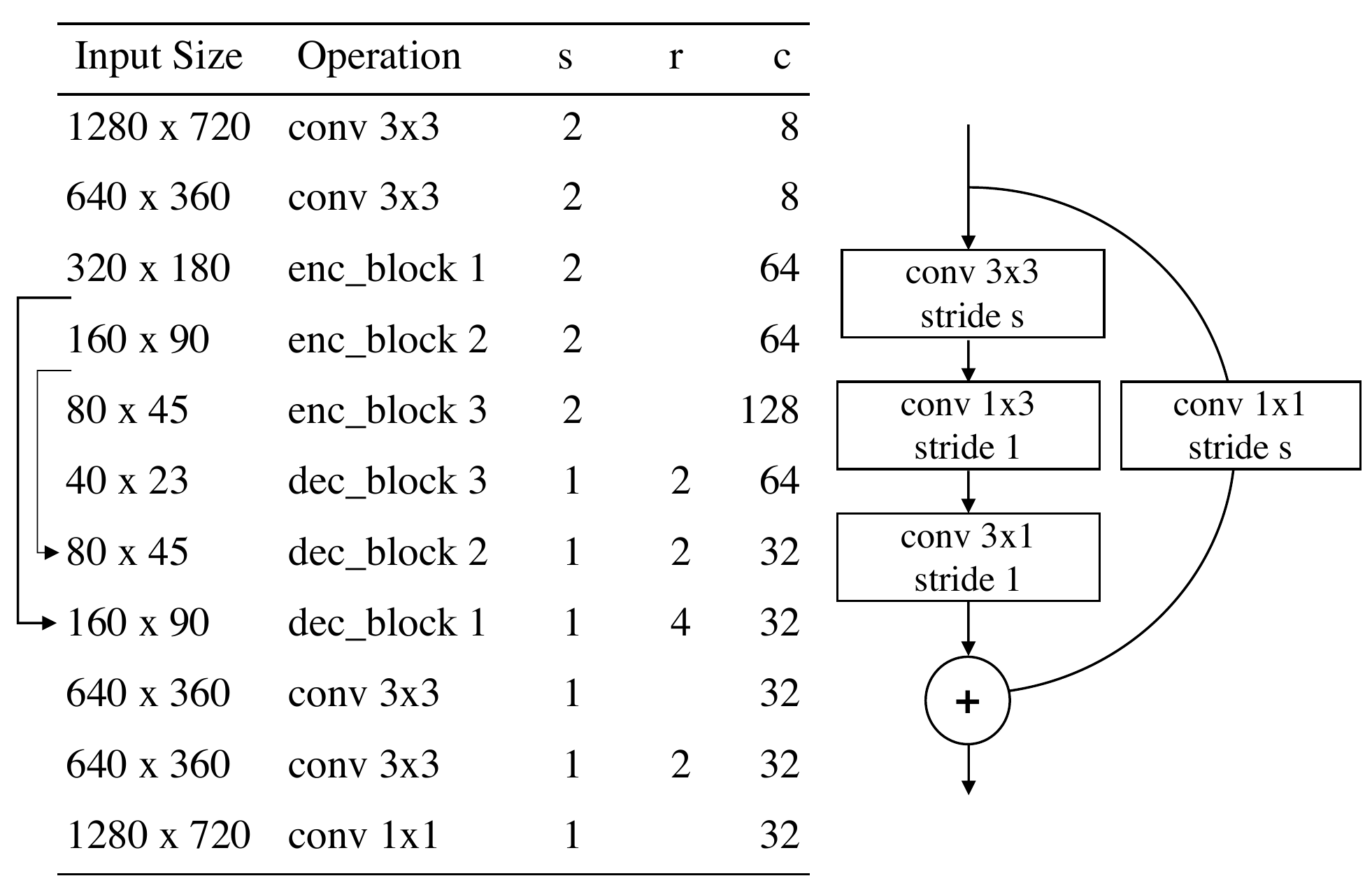}
\vspace{-2em}

\caption{Left: \jitnet\ architecture. Right: encoder/decoder block details.
         s = stride, r = resize, c = output channels.}
\label{fig:architecture}
\vspace{-1em}
\end{figure}

Our \jitnet\ architecture is a compact
encoder-decoder\,\cite{badrinarayanan2017segnet} composed of three modified
ResNet\,\cite{resnet} blocks. To reduce computation, we replace the second
3$\times$3 filter in each block with a separable filter (1$\times$3 followed by
a 3$\times$1) and also limit the number of channels for high resolution feature
maps.  To ensure fast training, we add skip connections from each encoder block
to the corresponding decoder block. This allows the gradient signal to propagate
efficiently to lower layers. We include diagnostic experiments to evaluate the
impact of these skip connections in supplemental.




Table~\ref{tab:network_costs} gives the parameter count, number of
floating-point operations, and runtime of both \jitnet\ and MRCNN on a frame of
720p video on an NVIDIA V100 GPU.  (We provide both inference and training costs
for \jitnet.) Compact segmentation models, such as those based on MobileNet
V2\,\cite{sandler2018inverted, tfmodelzoo}, are 3-4$\times$ slower than \jitnet\
at high resolution and are not designed for fast, stable online training.
We evaluate the MobileNet V2 architecture as the student model and
demonstrate that online distillation is viable using off-the-shelf
architectures. However, we find that \jitnet\ is more suitable for achieving
both higher accuracy and efficency. We also evaluate \jitnet\ variants on
standard semantic segmentation to ground it relative to other efficiency-oriented architectures. 
Both studies are included in the supplemental.

\addtolength{\tabcolsep}{-3pt}
\begin{table}[t!] \centering
    \footnotesize
    \begin{tabular}{l >{\raggedleft}p{1.1cm} >{\raggedleft}p{0.8cm} r
        >{\raggedleft}p{0.8cm} >{\raggedleft\arraybackslash}p{0.8cm}}\toprule
        Model & \multicolumn{2}{c}{FLOPS (B)} & Params (M) & \multicolumn{2}{c}{Time (ms)}\\
             \cmidrule{2-3}
             \cmidrule{5-6}
              & Infer & Train & & Infer & Train \\
        \midrule
        \jitnet\ & 15.2 & 42.0 & 3  & 7 & 30 \\
        MRCNN & 1390.0 & - & 141 & 300 & - \\
    \bottomrule
    \end{tabular}
    \vspace{-0.5em}
    \caption{ FLOPS (inference, training), parameter count, and runtime for both
    \jitnet\ and MRCNN. \jitnet\ has 47$\times$ fewer parameters and requires
    91$\times$ (inference) and 34$\times$ (training) fewer FLOPS than MRCNN
    inference.}     
    \label{tab:network_costs}
    \vspace{-1em}
\end{table}
\addtolength{\tabcolsep}{3pt}

\subsection{Online Training with Gradient Descent}
\label{sec:online_training} Online training presents many challenges:
training samples (frames) from the video stream are highly correlated,
there is continuous distribution shift in content (the past may not be representative of the future),
and teacher predictions used as a proxy for ``ground truth" at training
can exhibit temporal instability or errors. The method for updating \jitnet\
parameters must account for these challenges.

To generate target labels for training, we use the instance masks provided by
MRCNN above a confidence threshold, and convert them to pixel-level semantic
segmentation labels. All pixels where no instances are reported are labeled as
background. On most video streams, this results in a significantly higher
fraction of background compared to other classes. This imbalance reduces the
ability of the student model to learn quickly, especially for small objects, due
to most of the loss being weighted on background. We mitigate this issue by
weighting the pixel loss in each predicted instance bounding box (dilated by
15\%) five times higher than pixels outside boxes. This weighting focuses
training on the challenging regions near object boundaries and on small
objects. With these weighted labels, we compute the gradients for updating the
model parameters using weighted cross-entropy loss and gradient descent. Since
training \jitnet\ on a video from a random initialization would require
significant training to adapt to the stream, we pretrain \jitnet\ on the COCO
dataset, then adapt the pretrained model to each stream.


When fine-tuning models offline, it is common to only update a few layers or use
small learning rates to avoid catastrophic forgetting. In contrast, for online
adaptation, the goal is to minimize the cost of adapting the \jitnet\ model so
that it maintains high accuracy for current and near future video content.
Rapidly specializing the compact \jitnet\ to the temporal context retains high
accuracy at low-cost.  \emph{Therefore, we update all layers with high learning
rates.} Empirically, we find that gradient descent with high momentum (0.9) and
learning rate (0.01) works remarkably well for updating \jitnet\ parameters.  We
believe high momentum stabilizes training due to resilience to teacher
prediction noise. \textit{We use the same parameters for all online training
experiments.}

\vspace{-1em}
\begin{algorithm}
\setstretch{0.90}
\SetAlgoLined
\DontPrintSemicolon 
\SetKwInput{KwInput}{Input}
\SetKw{True}{true}
\SetKw{False}{false}
\SetKw{And}{and}
\SetKwInput{KwOutput}{Output}
\KwInput{$S_{0\dots n}$, ${u}_{max}$, ${\delta}_{min}$, ${\delta}_{max}$, ${a}_{thresh}$, $\theta_{0}$}
    \KwOutput{$P_{0\dots n}$}
$\delta$ $\leftarrow$ ${\delta}_{min}$ \;
\For{$t$ $\leftarrow$ 0 \KwTo n } {
    \eIf{ $t \equiv 0 \pmod{\delta}$} {
        $L_t$ $\leftarrow$ MaskRCNN($S_t$) \;
        $u$ $\leftarrow$ 0, $update$ $\leftarrow$ \True\;
        \While {$update$} {
            $P_t$ $\leftarrow$ JITNet($\theta_{t}$, $S_t$) \;
            $a_{curr}$ $\leftarrow$ MeanIoU($L_t$, $P_t$) \;
            \eIf {$u < u_{max}$ \And $a_{curr} < {a}_{thresh}$} {
                $\theta_{t}$ $\leftarrow$ UpdateJITNet($\theta_{t}$, $P_t$, $L_t$) \;
            }{ $update$ $\leftarrow$ \False \; }
            $u$ $\leftarrow$  $u + 1$ \;
        }
        \eIf {$a_{curr} > a_{thresh}$} {
            $\delta$ $\leftarrow$ min($\delta_{max}$, $2 \delta$)
        } {
            $\delta$ $\leftarrow$ max($\delta_{min}$, $\delta / 2$)
        }
    } {
        $P_t$ $\leftarrow$ JITNet($\theta_{t}$, $S_t$) \;
    }
    $\theta_{t+1}$ $\leftarrow$ $\theta_{t}$ \;
}
%
\caption{Online distillation}
\label{alg:online_distillation}
\end{algorithm}
\vspace{-1.5em}

\subsection{Adaptive Online Distillation}

Finally, we need to determine {\em when} the student needs supervision from the
teacher. One option is to run the teacher at a fixed rate (e.g., once every $n$
frames).  However, greater efficiency is possible using a dynamic approach that
adapts \jitnet\ with teacher supervision only when its accuracy drops.
Therefore, we require an algorithm that dynamically determines when it is
necessary to adapt \jitnet\ without incurring the cost of running the teacher
each frame to assess \jitnet's accuracy.


Our strategy is to leverage the teacher labels on prior frames not only for
training, but also for {\em validation}: our approach ramps up (or down) the
rate of teacher supervision based on recent student accuracy. Specifically, we
make use of exponential back-off\,\cite{goodman1988stability}, as outlined in
Algorithm~\ref{alg:online_distillation}. Inputs to our online distillation
algorithm are the video stream ($S_t$), maximum number of learning steps
performed on a single frame ($u_{max}$), the minimum/maximum frame strides
between teacher invocations ($\delta_{min}$, $\delta_{max}$), a desired accuracy
threshold ($a_{thresh}$), and the initial \jitnet\ model parameters
($\theta_0$). 

The algorithm operates in a streaming fashion and processes the frames in the
video in temporal order. The teacher is only executed on frames which are
multiples of the current stride ($\delta$). When the teacher is run, the
algorithm computes the accuracy of the current \jitnet\ predictions ($P_t$) with
respect to the teacher predictions ($L_t$).  If \jitnet\ accuracy is less than
the desired accuracy threshold (mean IoU), the model is updated using the
teacher predictions as detailed in the previous section. The \jitnet\ model is
trained until it either reaches the set accuracy threshold ($a_{thresh}$) or the
upper limit on update iterations ($u_{max}$) per frame. Once the training phase
ends, if \jitnet\ meets the accuracy threshold, the stride for running the
teacher is doubled; otherwise, it is halved (bounded by minimum and maximum
stride). The accuracy threshold is the only user-exposed knob in the algorithm.
As demonstrated in our evaluation, modifying the threshold's value allows for a
range of accuracy vs. efficiency trade-offs.

Even when consecutive video frames contain significant motion, their overall
appearance may not change significantly. Therefore, it is better to perform more
learning iterations on the current frame than to incur the high cost of running
the teacher on a new, but visually similar, frame. The maximum stride was chosen
so that the system can respond to changes within seconds (64~frames is about
2.6~seconds on 25~fps video). The maximum updates per frame is roughly the ratio
of \jitnet\ training time to teacher inference cost. We set $\delta_{min}$ and
$\delta_{max}$ to 8 and 64 respectively, and $u_{max}$ to 8 for all experiments.
\textit{We include further discussion and an ablation study of these parameters,
choices in network design, and training method in supplemental.}



%% file: dataset.tex
\section{Long Video Streams (LVS) Dataset}
\label{sec:dataset}

\renewcommand{\arraystretch}{0.2}
\begin{figure*}[t!]\centering
    \small
    \begin{tabularx}{\textwidth}{cc}
    {MRCNN \qquad\qquad\qquad\qquad \jitnet\ 0.9} & {MRCNN \qquad\qquad\qquad\qquad
        \jitnet\ 0.9 } \\
    \includegraphics[width=3.3in]{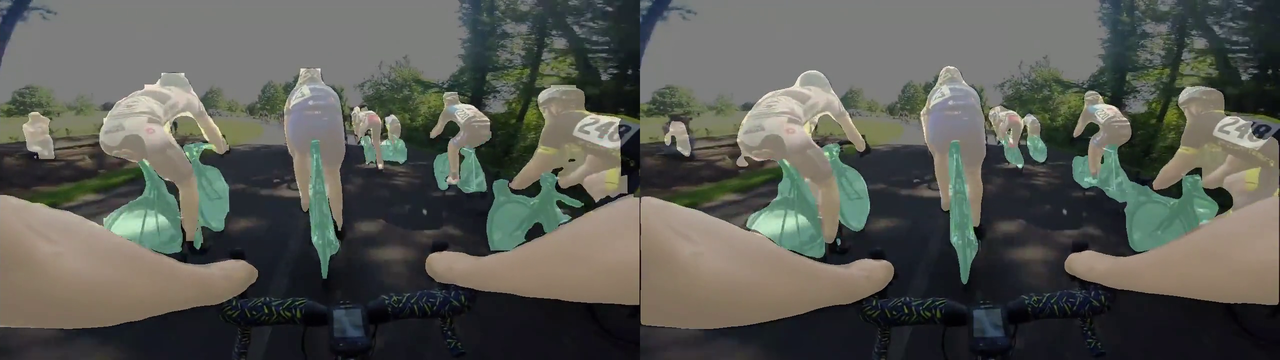} &
    \includegraphics[width=3.3in]{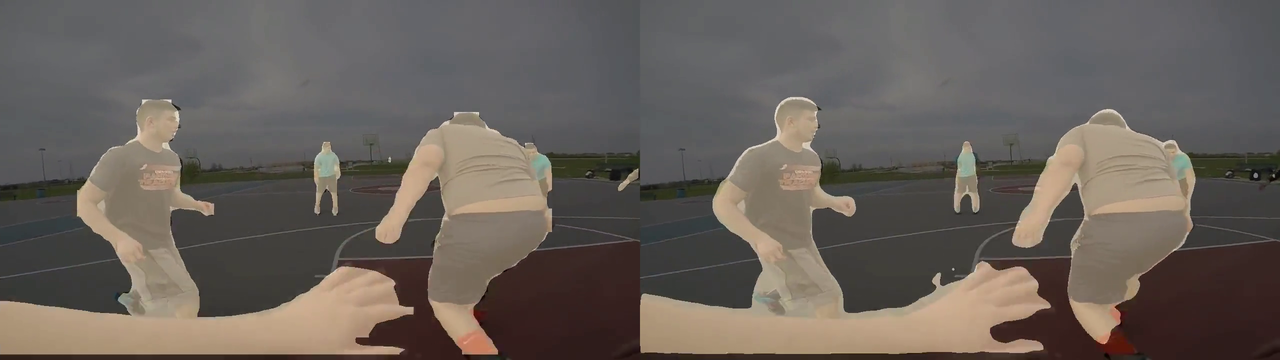} \\
    \includegraphics[width=3.3in]{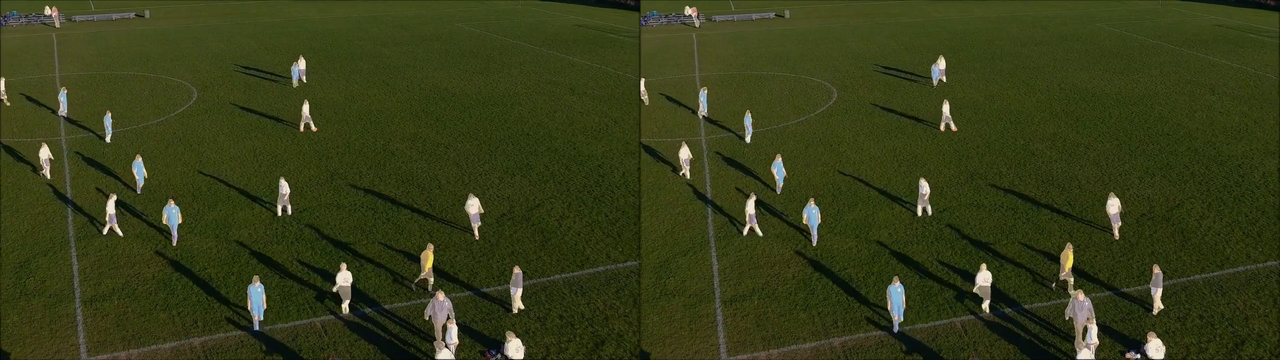}
    &
    \includegraphics[width=3.3in]{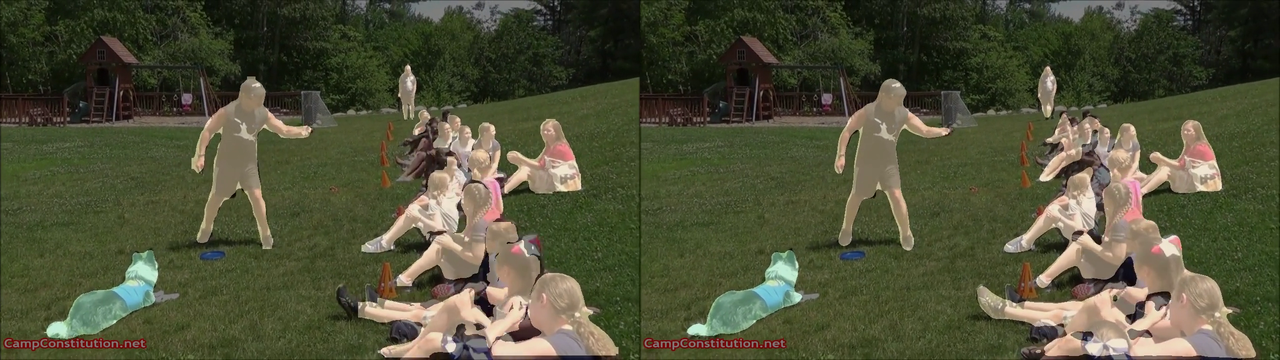} \\

    \includegraphics[width=3.3in]{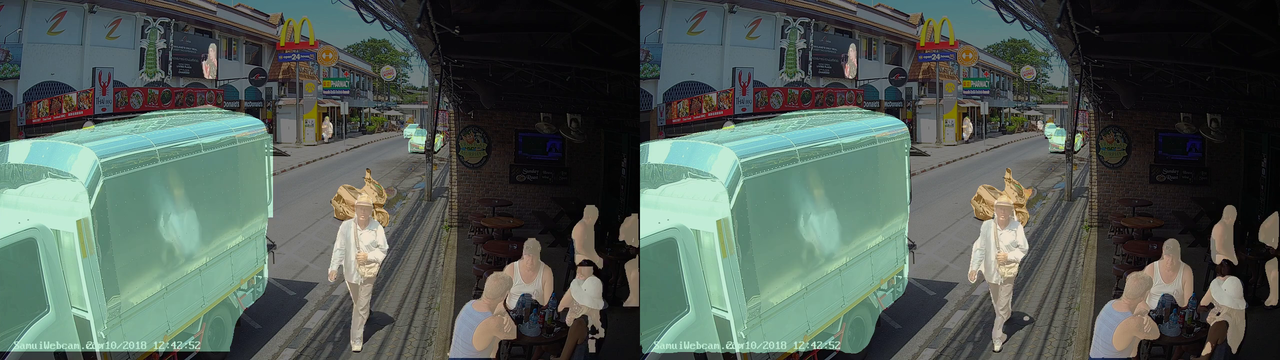} &
    \includegraphics[width=3.3in]{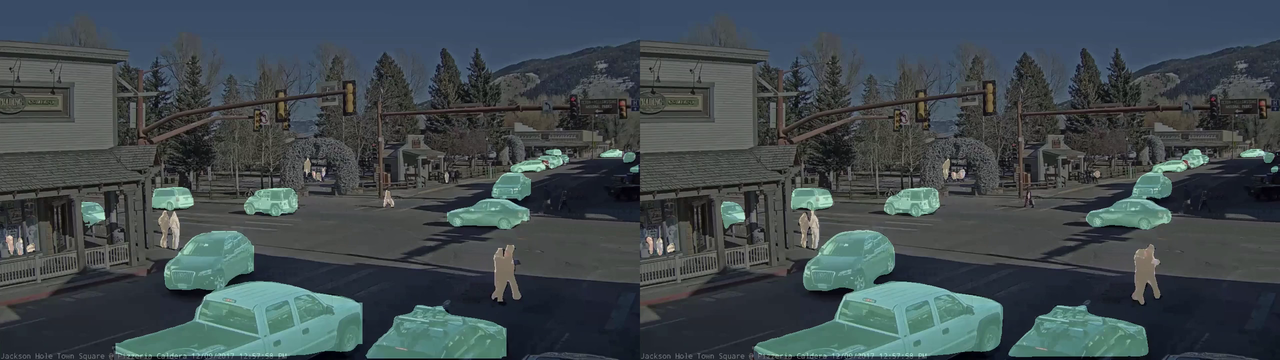}\\

\vspace{-0.5em}
    \end{tabularx} \caption{Frame segmentations generated by MRCNN (left) and \jitnet\ 0.9 (right)
                            from a subset of videos in the LVS dataset.}
\label{fig:scene_examples}
\vspace{-1.0em}
\end{figure*}


Evaluating fast video inference requires a dataset of long-running video streams
that is representative of real-world camera deployments, such as automatic retail checkout,
player analysis in sports,
traffic violation monitoring, and wearable device video analysis for augmented reality.
Existing large-scale video datasets have been designed to support training high-quality
models for various tasks, such as action detection\,\cite{kuehne2011hmdb,soomro2012ucf101},
object detection, tracking, and segmentation\,\cite{DAVIS,xu2018youtube}, and consist of carefully curated,
diverse sets of short video clips (seconds to a couple minutes).

We create a new dataset designed for evaluating techniques for efficient inference in real-world, long-running scenarios.
Our dataset, named the Long Video Streams dataset (LVS), contains 30~HD videos, each
30~minutes in duration and at least 720p~resolution. (900 minutes total; for comparison, YouTube-VOS~\cite{xu2018youtube} is 345 minutes.)
Unlike other datasets for efficient inference, which consist of streams from fixed-viewpoint cameras such as traffic cameras~\cite{Jiang:2018:chameleon},
we capture a diverse array of challenges: from fixed-viewpoint cameras, to constantly moving and zooming television cameras, and hand-held and egocentric video.
Given the nature of these video streams, the most commonly occurring objects include people, cars, and animals.

It is impractical to obtain ground truth, human-labeled segmentations for all
900 minutes (1.6 million frames) of the dataset.  Therefore, we curate a set of
representative videos and use MRCNN\,\cite{he2017mask} to generate predictions
on all the frames.  (We evaluated other segmentation models such as
DeepLab~V3\,\cite{chen2017rethinking} and Inplace~ABN\,\cite{inplaceabn}, and
found MRCNN to be produce the highest quality labels.) We use the
highest-quality MRCNN~\cite{detectron} without test-time data augmentation, and
provide its output for all dataset frames to aid evaluation of classification,
detection, and segmentation (semantic and instance level) methods.
Figure~\ref{fig:scene_examples} shows a sampling of videos from the dataset
with their corresponding MRCNN segmentations (left image in each group). We
refer readers to supplemental for additional dataset details and
visualizations of MRCNN predictions.

%% file: experiments.tex
\section{Evaluation}
\label{sec:evaluation}
To evaluate online distillation as a strategy for efficient video segmentation,
we compare its accuracy and cost with an alternative motion-based interpolation
method\,\cite{zhu2017deep} and an online approach for video object
segmentation~\cite{caelles2017one}. While our focus is evaluating accuracy and
efficiency on long video streams (LVS), we also include results on the DAVIS
video benchmark\,\cite{DAVIS} in supplemental.

\renewcommand{\arraystretch}{0.8}
\begin{table*}[t!]
    \footnotesize \centering
    \begin{tabular}{L{2.8cm}C{1.25cm}C{1.75cm}C{1.5cm}C{2.5cm}C{2.5cm}C{2.5cm}}\toprule
         &  Offline & \multicolumn{2}{c}{Flow\,\cite{zhu2017deep}} &
        \multicolumn{3}{c}{Online Distillation} \\
        \cmidrule{3-4} \cmidrule{5-7}
        Video &  Oracle & {Slow (2.2$\times$)} & {Fast (3.2$\times$)} & {\jitnet\ 0.7} &
        {\jitnet\ 0.8} & {\jitnet\ 0.9} \\
        & (20\%)       & (12.5\%) & ( 6.2\%) & & & \\
\arrayrulecolor{black!30}\midrule
{\textbf{Overall}}& 80.3& 76.6& 65.2& 75.5 (17.4$\times$, 3.2\%)& 78.6 (13.5$\times$, 4.7\%)& 82.5 ($\times$7.5, 8.4\%)\\
\arrayrulecolor{black!30}\midrule
         \multicolumn{7}{c}{Category Averages}\\
\arrayrulecolor{black!30}\midrule
{Sports (Fixed)}& 87.5& 81.2& 71.0& 80.8 (24.4$\times$, 1.6\%)& 82.8 (21.8$\times$, 1.8\%)& 87.6 (10.4$\times$, 5.1\%)\\
\arrayrulecolor{black!30}\midrule
{Sports (Moving)}& 82.2& 72.6& 59.8& 76.0 (20.6$\times$, 2.1\%)& 79.3 (14.5$\times$, 3.6\%)& 84.1 (6.0$\times$, 9.1\%)\\
\arrayrulecolor{black!30}\midrule
{Sports (Ego)}& 72.3& 69.4& 55.1& 65.0 (13.6$\times$, 3.7\%)& 70.2 (9.1$\times$, 6.0\%)& 75.0 (4.9$\times$, 10.4\%)\\
\arrayrulecolor{black!30}\midrule
{Animals}& 89.0& 83.2& 73.4& 82.9 (21.7$\times$, 1.9\%)& 84.3 (19.6$\times$, 2.2\%)& 87.6 (14.3$\times$, 4.4\%)\\
\arrayrulecolor{black!30}\midrule
{Traffic}& 82.3& 82.6& 74.0& 79.1 (11.8$\times$, 4.6\%)& 82.1 (8.5$\times$, 7.1\%)& 84.3 (5.4$\times$, 10.1\%)\\
\arrayrulecolor{black!30}\midrule
{Driving/Walking}& 50.6& 69.3& 55.9& 59.6 (5.8$\times$, 8.6\%)& 63.9 (4.9$\times$, 10.5\%)& 66.6 (4.3$\times$, 11.9\%)\\
\arrayrulecolor{black!30}\midrule
         \multicolumn{7}{c}{Subset of Individual Video Streams}\\
\arrayrulecolor{black!30}\midrule
{Table Tennis (P)}& 89.4& 84.8& 75.4& 81.5 (24.7$\times$, 1.6\%)& 83.5 (24.1$\times$, 1.6\%)& 88.3 (12.9$\times$, 3.4\%)\\
\arrayrulecolor{black!30}\midrule
{Kabaddi (P)}& 88.2& 78.9& 66.7& 83.8 (24.8$\times$, 1.6\%)& 84.5 (23.5$\times$, 1.7\%)& 87.9 (7.8$\times$, 6.3\%)\\
\arrayrulecolor{black!30}\midrule
{Figure Skating (P)}& 84.3& 54.8& 37.9& 72.3 (15.9$\times$, 2.8\%)& 76.0 (11.4$\times$, 4.1\%)& 83.5 (5.4$\times$, 9.4\%)\\
\arrayrulecolor{black!30}\midrule
{Drone (P)}& 74.5& 70.5& 58.5& 70.8 (15.4$\times$, 2.8\%)& 76.6 (6.9$\times$, 7.2\%)& 79.9 (4.1$\times$, 12.5\%)\\
\arrayrulecolor{black!30}\midrule
{Birds (Bi)}& 92.0& 80.0& 68.0& 85.3 (24.5$\times$, 1.6\%)& 85.7 (24.2$\times$, 1.6\%)& 87.9 (21.7$\times$, 1.8\%)\\
\arrayrulecolor{black!30}\midrule
{Dog (P,D,A)}& 86.1& 80.4& 71.1& 78.4 (19.0$\times$, 2.2\%)& 81.2 (13.8$\times$, 3.2\%)& 86.5 (6.0$\times$, 8.4\%)\\
\arrayrulecolor{black!30}\midrule
{Ego Dodgeball (P)}& 82.1& 75.5& 60.4& 74.3 (17.4$\times$, 2.5\%)& 79.5 (13.2$\times$, 3.4\%)& 84.2 (6.1$\times$, 8.2\%)\\
\arrayrulecolor{black!30}\midrule
{Biking (P,Bk)}& 70.7& 71.6& 61.3& 68.2 (12.7$\times$, 3.5\%)& 72.3 (6.7$\times$, 7.3\%)& 75.3 (4.1$\times$, 12.4\%)\\
\arrayrulecolor{black!30}\midrule
{Samui Street (P,A,Bk)}& 80.6& 83.8& 76.5& 78.8 (8.8$\times$, 5.5\%)& 82.6 (5.3$\times$, 9.5\%)& 83.7 (4.2$\times$, 12.2\%)\\
\arrayrulecolor{black!30}\midrule
{Driving (P,A,Bk)}& 51.1& 72.2& 59.7& 63.8 (5.7$\times$, 8.8\%)& 68.2 (4.5$\times$, 11.5\%)& 66.7 (4.1$\times$, 12.4\%)\\
\arrayrulecolor{black}\bottomrule
    \end{tabular}
    \vspace{-0.5em}
    \caption{Comparison of accuracy (mean IoU over all the classes excluding
    background), runtime speedup relative to MRCNN (where applicable), and the
    fraction of frames where MRCNN is executed.  Classes
    present in each video are denoted by letters (A - Auto, Bi - Bird, Bk - Bike, D -
    Dog, E - Elephant, G - Giraffe, H - Horse, P - Person).  Overall, online distillation using
    \jitnet\ provides a better accuracy/efficiency tradeoff than baseline
    flow based methods\,\cite{zhu2017deep} and has accuracy comparable to
    oracle offline models.} 
    \label{tab:main_table}
    \vspace{-1em}
\end{table*}

\subsection{Experimental Setup}
Our evaluation focuses on both the efficiency and accuracy of semantic
segmentation methods relative to MRCNN. Although MRCNN trained on the COCO
dataset can segment 80 classes, LVS video streams captured from a single camera
over a span of 30 minutes typically encounter a small subset of these classes.
For example, none of the indoor object classes such as appliances and cutlery
appear in outdoor traffic intersection or sports streams. Therefore, we measure
accuracy only on classes which are present in the stream and have reliable MRCNN
predictions. Our evaluation focuses on object classes which can independently
move, since stationary objects can be handled efficiently using simpler methods.
We observed that MRCNN often confused if an instance is a car, truck, or a bus, so
to improve temporal stability we combine these classes into a single class
``auto" for both training and evaluation.  Therefore, we only evaluate accuracy
on the following classes: bird, bike, auto, dog, elephant, giraffe, horse, and
person. Table~\ref{tab:main_table} shows the classes that are evaluated in each
individual stream as an abbreviated list following the stream name.

All evaluated methods generate pixel-level predictions for each class in the
video. We use mean intersection over union (mean IoU) over the classes in each
video as the accuracy metric. All results are reported on the first 30,000
frames of each video ($\approx$16-20 minutes due to varying fps) unless otherwise specified.
Timing measurements for \jitnet, MRCNN (see Table~\ref{tab:network_costs}),
and other baseline methods are performed using TensorFlow~1.10.1
(CUDA~9.2/cuDNN~7.3) and PyTorch~0.4.1 for MRCNN on an NVIDIA~V100 GPU. All
speedup numbers are reported relative to wall-clock time of MRCNN. Note
that MRCNN performs instance segmentation whereas \jitnet\ performs semantic
segmentation on a subset of classes.

\begin{figure*}[t!]\centering
\renewcommand{\arraystretch}{0.1}
\setlength{\tabcolsep}{0pt}
    \tiny
    \begin{tabularx}{\textwidth}{ccc}
        \includegraphics[width=0.33\textwidth]{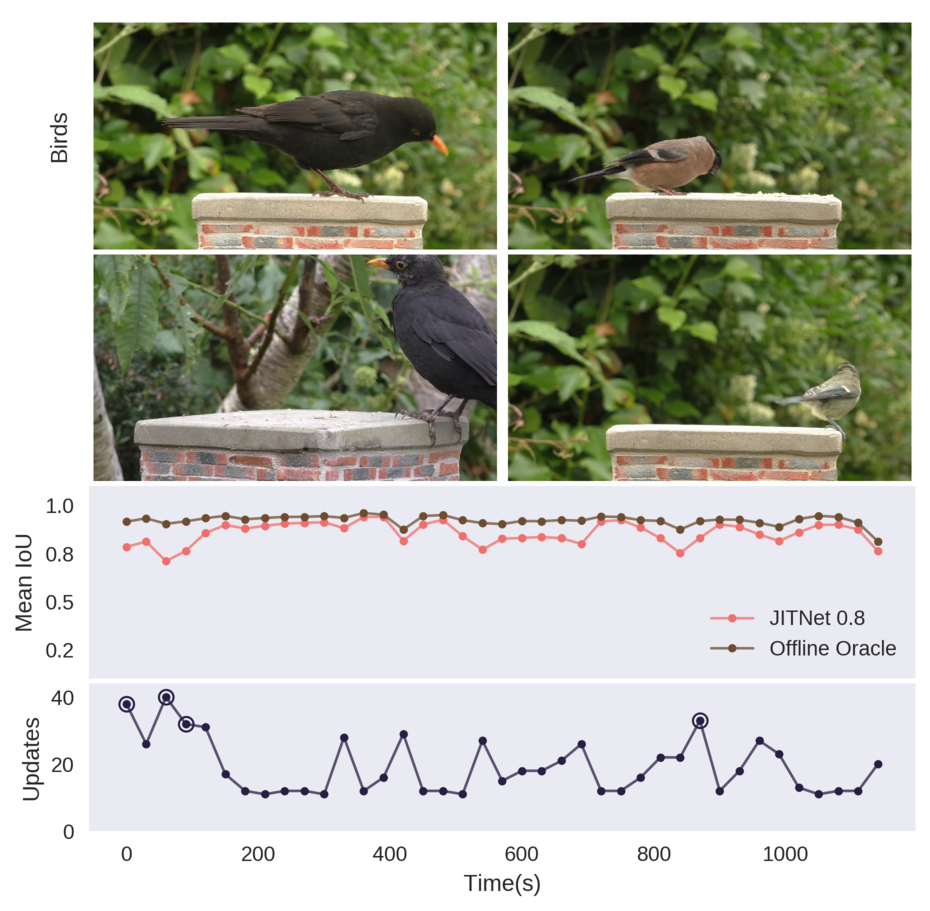} &
        \includegraphics[width=0.33\textwidth]{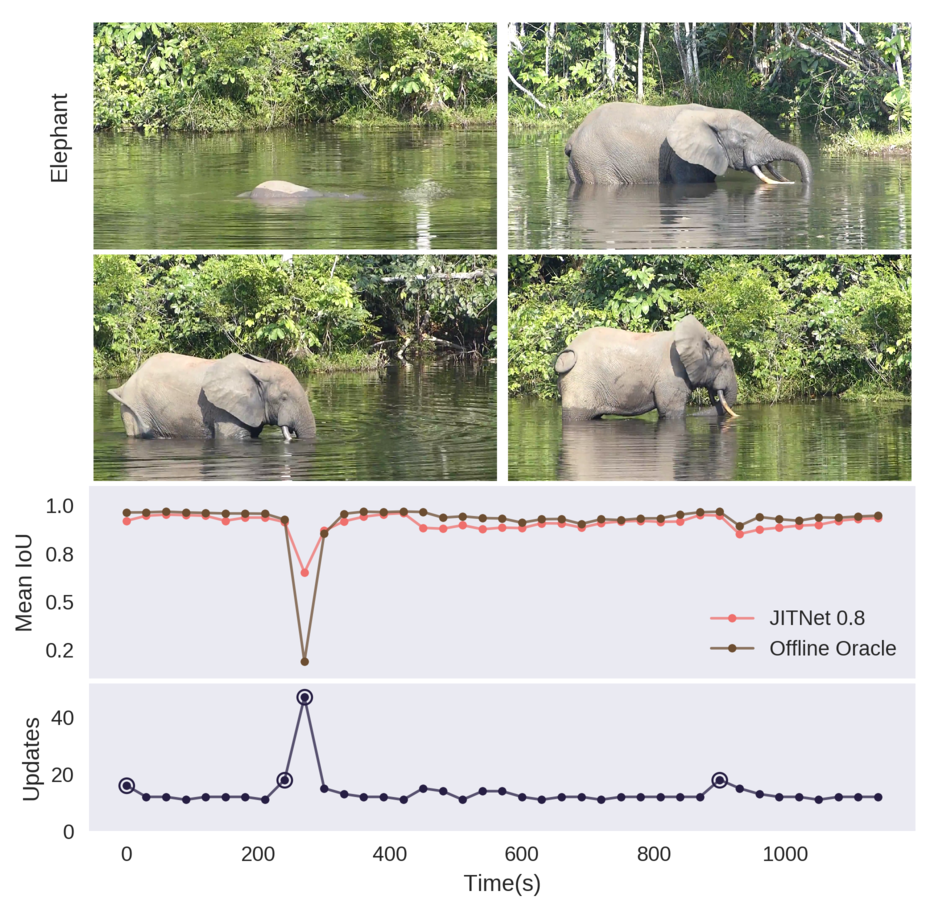} &
        \includegraphics[width=0.33\textwidth]{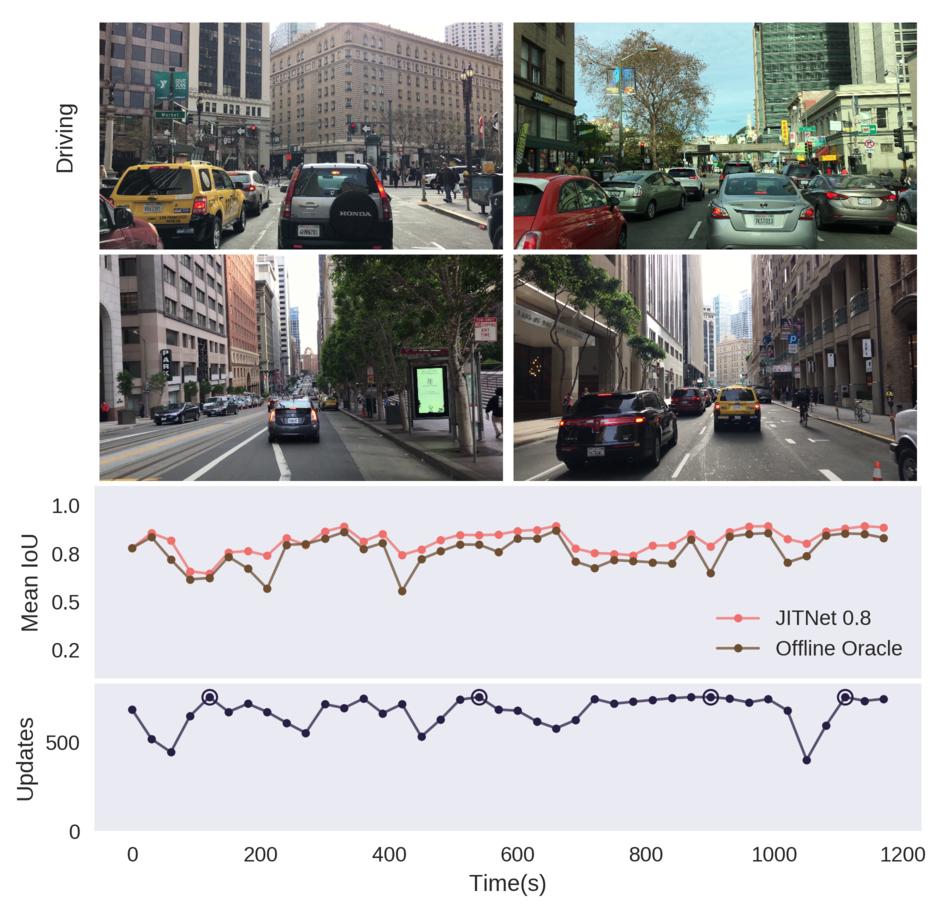} \\
 \end{tabularx}
    \vspace{-1.2em}
\caption{Top graph: the accuracy of \jitnet~0.8 and Offline
    Oracle relative to MRCNN. Bottom graph: the number of updates to \jitnet\ during online
    distillation.  Plotted points are averages over a 30~second interval of the video. Images correspond to circled points in bottom plot, and show times where \jitnet\ required frequent training to maintain accuracy.}
    \vspace{-1.4em}
\label{fig:time_scenes}
\end{figure*}

\subsection{Accuracy vs. Efficiency of Online Distillation}
\label{sec:accuracyEval}
Table~\ref{tab:main_table} gives the accuracy and performance of online
distillation using \jitnet\ at three different accuracy thresholds:
\jitnet~0.7, 0.8, and 0.9. Performance is the average speedup
relative to MRCNN runtime, \emph{including the cost of teacher
evaluation and online \jitnet\ training}. To provide intuition on the
speedups possible on different types of videos, we organize LVS into
categories of similar videos and show averages
for each category (e.g., Sports (Moving) displays average results for
seven sports videos filmed with a moving camera), as well as provide per-video
results for a selection of 10 videos.
We also show the fraction of frames for which MRCNN predictions
are used. For instance, on the Kabaddi video stream, \jitnet~0.8 is 23.5 times
faster than MRCNN, with a mean IoU of 84.5, and uses 510 frames
out of 30,000 (1.7\%) for supervision. Detailed results and videos for all streams,
showing both MRCNN and JITNet predictions side-by-side for qualitative
comparison, are provided in supplemental.

On average, across all sequences, \jitnet~0.9 maintains 82.5 mean IoU with
7.5$\times$ runtime speedup (11.3$\times$ in FLOPs). In the lower accuracy regime, \jitnet~0.7 is 17.4$\times$
faster on average (26.2$\times$ in FLOPs) while maintaining a mean IoU of 75.5.
Mean IoUs in the table exclude the background class, where all the methods have
high accuracy. As expected, when the accuracy threshold is increased, \jitnet\
improves in accuracy but uses a larger fraction of teacher frames for
supervision. Average speedup on sports streams from fixed cameras is higher than
that for moving cameras. Even on challenging egocentric sports videos with
significant motion blur, \jitnet~0.9 provides 4.9$\times$ speedup while
maintaining 75.0 mean IoU.

Although \jitnet\ accuracy on the Sports~(Fixed), Sports~(Moving), Animals, and Traffic
categories suggests potential for improvement, we observe that for streams with
large objects, it is often difficult to qualitatively discern if \jitnet\ or
MRCNN produces higher quality predictions. Figure~\ref{fig:scene_examples}
displays sample frames with both MRCNN (left) and \jitnet\ (right) predictions
(zoom in to view details). The boundaries produced by \jitnet\ on large objects
(1st row) are smoother than MRCNN, since MRCNN
generates low-resolution masks (28 $\times$ 28) that are upsampled to full
resolution. However, for videos containing small objects, such as traffic camera
(Figure~\ref{fig:scene_examples}, 3rd row, right) or aerial views
(2nd row, left), MRCNN produces sharper
segmentations. \jitnet's architecture and operating resolution would need to be
improved to match MRCNN segmentations on small objects.

Streams from the Sports (Ego) category exhibit significant motion blur due to
fast motion. Teacher predictions on blurred frames can be unreliable and
lead to disruptive model updates. The Driving/Walking streams traverse a busy
downtown and a crowded beach, and are expected to be challenging for online
distillation since object instances persist on screen for only short
intervals in these videos. Handling these scenarios more accurately would require
faster methods for online model adaptation.

\label{sec:interpolation}
\begin{figure}[t!]\centering
    \includegraphics[width=0.45\textwidth]{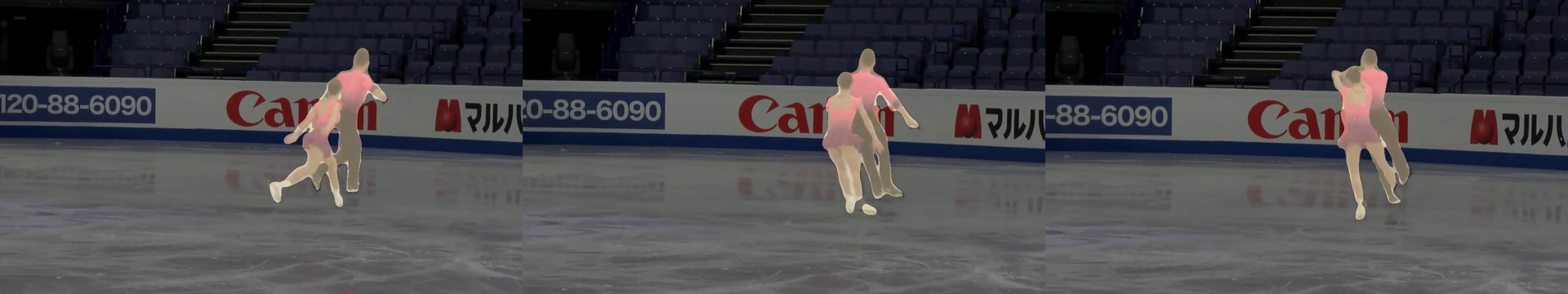}\\
    \includegraphics[width=0.45\textwidth]{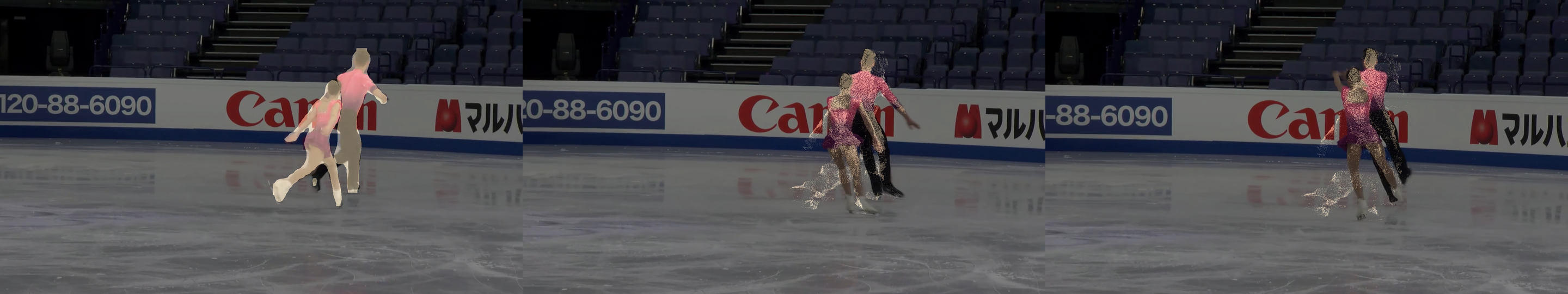}
    \caption{Top: \jitnet~0.9 predictions on a sequence of three frames which
    are roughly 0.13 seconds apart (4 frames apart) in the Figure Skating video.
    Bottom: Large deformations, object and camera motion prove challenging to
    flow based interpolation.}
    \vspace{-1em}
\label{fig:skating_deform}
\vspace{-0.6em}
\end{figure}

\subsection{Comparison with Offline Oracle Specialization}
The prior section shows that a \jitnet\ model pre-trained only on COCO can be
continuously adapted to a new video stream with only
modest online training cost. We also compare the accuracy
of just-in-time adaptation to the results of
specializing \jitnet\ to the contents of the each stream \textit{entirely offline}, and performing no online training.
To simulate the effects of near best-case offline pre-training, we train \jitnet\
models on every 5th frame of the entire 20~minute test video sequence (6,000 training frames).
We refer to these models as ``offline oracle" models since they are constructed
by pre-training on the test set, and serve as a strong baseline for the accuracy
achievable via offline specialization.  All offline oracle models were
pre-trained on COCO, and undergo \textit{one hour} of pre-training on 4 GPUs using
traditional random-batch SGD. (See supplemental for further
details.) Recall that in contrast, \emph{online adaptation incurs no
pre-training cost} and trains in a streaming fashion.

As shown in Table~\ref{tab:main_table}, \jitnet~0.9 is on average more accurate
than the offline oracle. Note that \jitnet~0.9 uses only 8.4\% of frames on average
for supervision, while the oracle is trained using 20\%. This trend also holds for the
subcategory averages. This suggests that the compact
\jitnet\ model does not have sufficient capacity to fully capture the diversity present in the
20~minute stream.

Figure~\ref{fig:time_scenes} shows mean IoU of \jitnet~0.8 and the offline
oracle across time for three videos. The top plot displays mean IoU of both
methods (data points are averages over 30~second time intervals).
The bottom plot displays the number of \jitnet\ model updates in each interval. Images above the
plots are representative frames from time intervals requiring the most
\jitnet\ updates. In the Birds video (left), these intervals correspond to events when new birds appear.
In comparison, the Elephant video (center)
contains a single elephant from different viewpoints and camera angles.
The offline oracle model incurs a significant accuracy drop when the elephant dips into water.
(This rare event makes up only a small fraction of the offline training set.) 
\jitnet~0.8 displays a smaller drop since it specializes immediately to the novel scene characteristics.
The Driving video (right) is challenging for both the offline oracle and online \jitnet\ since it features significant visual diversity and continuous change.
However, while the mean IOU of both methods is lower, online
adaptation consistently outperforms the offline oracle in this case as well.

\subsection{Comparison with Motion-Based Interpolation}
An alternative approach to improving segmentation efficiency on video is to
compute teacher predictions on a sparse set of frames and interpolate the results using flow.
Table~\ref{tab:main_table} shows two baselines that propagate pixel segmentations using Dense Feature Flow\,\cite{zhu2017deep}, although we upgrade the flow estimation network from FlowNet2\,\cite{ilg2017flownet} to modern methods. (We propagate labels,
not features, since this was shown to be as effective\,\cite{zhu2017deep}.) The
expensive variant (Flow (Slow)) runs MRCNN every 8th frame and uses PWC-Net\,\cite{sun2018pwcnet} to estimate optical flow between frames. MRCNN labels are propagated to the
next seven frames using the estimated flow. The fast variant (Flow
(Fast)) uses the same propagation mechanism but runs MRCNN every 16th frame and
uses a faster PWC-Net. Overall \jitnet~0.7 is 2.8$\times$ faster and more accurate than
the fast flow variant, and \jitnet~0.9 has significantly higher accuracy
than the slow flow variant except in the Driving/Walking category.

Figure~\ref{fig:skating_deform} illustrates the challenge of using flow to
interpolate sparse predictions. Notice how the ice skaters in the video undergo
significant deformation, making them hard to track via flow.  In contrast,
online distillation trains \jitnet\ to learn the appearance of scene objects (it
leverages temporal coherence by reusing the model over local time windows),
allowing it to produce high-quality segmentations despite complex motion.
The slower flow baseline performs well compared to online adaptation on rare classes
in the Driving (Bike) and Walking (Auto) streams, since flow is
agnostic to semantic classes. Given the orthogonal nature of flow and online
adaptation, it is possible a combination of these approaches could be used to
handle streams with rapid appearance shifts.

\renewcommand{\arraystretch}{0.8}
\begin{table}
    \small\centering
    \begin{tabular}{L{2.5cm}C{0.75cm}C{0.75cm}C{2.5cm}}
        \toprule
        Category & \multicolumn{2}{c}{OSVOS (3.3\%)} & {\jitnet\ 0.8} \\
                   \cmidrule{2-3}
                 & A & B &  \\
\arrayrulecolor{black!30}\midrule
{\textbf{Overall}}& 59.9& 60.0& 77.4 (14.5$\times$, 4.6\%)\\
\arrayrulecolor{black!30}\midrule
{Sports (Fixed)}& 75.7& 75.7& 82.3 (24.0$\times$, 1.6\%)\\
\arrayrulecolor{black!30}\midrule
{Sports (Moving)}& 69.1& 69.3& 78.7 (16.3$\times$, 2.9\%)\\
\arrayrulecolor{black!30}\midrule
{Sports (Ego)}& 67.6& 68.1& 74.8 (9.5$\times$, 5.9\%)\\
\arrayrulecolor{black!30}\midrule
{Animals}& 79.3& 79.8& 86.0 (19.7$\times$, 2.1\%)\\
\arrayrulecolor{black!30}\midrule
{Traffic}& 22.3& 21.9& 70.8 (8.4$\times$, 7.7\%)\\
\arrayrulecolor{black!30}\midrule
{Driving/Walking}& 36.7& 36.3& 66.8 (4.3$\times$, 11.8\%)\\
\arrayrulecolor{black}\bottomrule
   \end{tabular}
    \vspace{-0.5em}
    \caption{\jitnet~0.8 generates higher accuracy segmentations than OSVOS on LVS and is two orders of magnitude lower cost. Percentages give the fraction of frames used for MRCNN supervision.}
    \vspace{-1.4em}
    \label{tab:osvos}
\end{table}

\subsection{Comparison with Video Object Segmentation}

Although not motivated by efficiency, video object segmentation (VOS)
solutions employ a form of online adaptation: they train a model to segment future video frames
 based on supervision provided in the first frame. We evaluate the accuracy of the OSVOS\,\cite{caelles2017one}
approach against \jitnet\ on two-minute segments of each LVS video.  (OSVOS was too expensive to run on longer segments.)
For each 30-frame interval of the segment, we use MRCNN to generate a starting foreground mask, train the OSVOS model on the starting mask, and use the resulting model for segmenting the next 29 frames.
We train OSVOS for 30~seconds on each starting frame, which requires approximately {\em one hour} to run OSVOS on each two-minute video segment.
Since segmenting all classes in the LVS videos would require running OSVOS once per class,
we  run OSVOS on only one class per video (person or animal class in each stream)
and compare \jitnet\ accuracy with OSVOS on the designated class. (Recall \jitnet\ segments all classes.)
Furthermore, we run two configurations of OSVOS: in mode (A) we use the OSVOS model from the previous 30-frame interval as the starting point for training in the next interval (a form of continuous adaptation). In mode (B) we reset to the pre-trained OSVOS model for each 30-frame interval.

Table~\ref{tab:osvos} compares the accuracy of both OSVOS variants to online distillation with \jitnet.  The table also provides model accuracy, runtime speedup relative to MRCNN, and the fraction of frames used by \jitnet~0.8 for supervision in the two-minute interval. Overall \jitnet~0.8 is more accurate than OSVOS and {\em two orders of magnitude} faster. On Traffic streams, which have small objects, and Driving/Walking streams with rapid appearance changes, OSVOS has significantly lower
accuracy than \jitnet~0.8. We also observe that the mode A variant of OSVOS (continuously adapted) performs worse than the variant which is re-initialized. We believe the \jitnet\ architecture could be employed as a means to significantly accelerate online VOS methods like OnAVOS\,\cite{voigtlaender:2017:onAVOS} or more recent OSVOS-S\,\cite{osvoss} (uses MRCNN predictions every frame).

%% file: conclusion.tex
\section{Conclusion}

In this work we demonstrate that for common, real-world video streaming
scenarios, it is possible to perform online distillation of compact (low cost)
models to obtain semantic segmentation accuracy that is comparable with an
expensive high capacity teacher. Going forward, we hope that our results
encourage exploration of online distillation for domain adaptation and self-supervised learning. More generally,
with continuous capture of high-resolution video streams becoming increasingly
commonplace, we believe it is relevant for the broader community to think about
the design and training of models that are not trained offline on carefully
curated datasets, but instead continuously evolve each day with the data that
they observe from specific video streams. We hope that the Long Video Streams
dataset serves this line of research.

%% file: ack.tex
\\
\textbf{Acknowledgement} This research is based upon work supported in part by
NSF Grant 1618903 and IIS-1422767, the Intel Science and Technology Center for
Visual Cloud Systems (ISTC-VCS), the Office of the Director of National
Intelligence (ODNI), Intelligence Advanced Research Projects Activity (IARPA),
via IARPA R \& D Contract No. D17PC00345, and the Defense Advanced Research
Projects Agency (DARPA) under Contract No. HR001117C0051, and a Google Faculty
Fellowship. The views and conclusions contained herein are those of the authors
and should not be interpreted as necessarily representing the official policies
or endorsements, either expressed or implied, of ODNI, IARPA, or the U.S.
Government. The U.S. Government is authorized to reproduce and distribute
reprints for Governmental purposes notwithstanding any copyright annotation
thereon. 

%% file: supp_arxiv.tex
\section{Supplementary}

\subsection{Qualitative Comparison Videos}
We include one minute video clips (videos.zip) from a selection of video streams in the LVS
dataset, with Mask R-CNN and \jitnet\ 0.9 predictions overlaid on the left and
right respectively. All videos are subsampled by 4$\times$ temporally to reduce
file size. Full videos can be found at this anonymous YouTube channel:
\url{https://www.youtube.com/channel/UC-T0FerolHcDDKs2BZQEmrQ}

\subsection{Online Distillation Ablation Study}

\begin{table*}[t!]
    \footnotesize \centering
    \begin{tabular}{L{2.2cm}C{1.5cm}C{0.58cm}C{0.58cm}C{0.58cm}C{0.58cm}C{0.58cm}C{0.58cm}C{0.58cm}C{0.58cm}C{0.58cm}C{0.58cm}C{0.58cm}C{0.58cm}C{0.58cm}}\toprule
        &  \multicolumn{12}{c}{JITNet} & \multicolumn{2}{c}{MobileNet}\\
        \cmidrule(lr){2-13} \cmidrule(lr){14-15} 
         &  Baseline & \multicolumn{2}{c}{Max Updates} &
        \multicolumn{2}{c}{Learning Rate} & \multicolumn{2}{c}{Min Stride} & \multicolumn{2}{c}{Width} & Skip & \multicolumn{2}{c}{Scale} & \multicolumn{2}{c}{Output Stride}\\
        \cmidrule(lr){3-4} \cmidrule(lr){5-6} \cmidrule(lr){7-8} \cmidrule(lr){9-10} \cmidrule(lr){12-13} \cmidrule(lr){14-15}
        & & 4 & 16 & 0.001 & 0.1 & 4 & 16 & 0.5 & 2.0 & No & 0.5 & 0.75 & 8 & 16 \\
    \arrayrulecolor{black}\midrule
        Accuracy (mIoU) & 78.7 & 77.3 & 78.0 & 75.6 & 16.7 & 79.8 & 76.1 & 62.0 & 80.3 & 76.0 & 75.9 & 78.1 & 75.3 & 74.6  \\
    \arrayrulecolor{black!30}\midrule
        Speed Up & 19.2$\times$ & 17.2$\times$ & 22.8$\times$ & 14.3$\times$ & 7.3$\times$ & 17.1$\times$ & 22.9$\times$ & 14.9$\times$ & 12.4$\times$ & 12.4$\times$ & 20.0$\times$ & 22.0$\times$ & 9.5$\times$ & 16.8$\times$ \\
    \arrayrulecolor{black!30}\midrule
        Teacher Samples & 5.0\% & 6.1\% &  3.7\% & 6.7\% & 10.6\% & 7.7\% & 3.3\% & 5.1\% & 4.2\% & 6.3\% & 6.2\% & 5.3\% & 5.0\% & 5.8\% \\
    \arrayrulecolor{black!30}\midrule
        Inference (FLOPS) &  \multicolumn{7}{c}{15.2} & 11.8 & 47.9 & 14.3 & 4.6 & 10.3 & 60.3 & 18.3 \\
    \arrayrulecolor{black!30}\midrule
        Training (FLOPS) & \multicolumn{7}{c}{42.0} & 31.8 & 140.4 & 39.4 & 8.6 & 22.4 & 176.1 & 53.0 \\
\arrayrulecolor{black}\bottomrule
    \end{tabular} \caption{Comparison of different input parameter settings to
    the online distillation algorithm. The algorithm is robust to all parameter
    changes except very high learning rates, where online training becomes
    unstable.}
    \label{tab:ablation_table}
\end{table*}

Our online distillation approach has several parameters (maximum updates
($u_{max}$),
minimum stride ($\delta_{min}$), learning rate and network size) that enable different trade-offs
between accuacy and efficiency. Here, we study the impact of these parameters on
the accuracy vs. efficiency trade-off on a subset of six video streams (which are
representative of different scenarios) in the LVS dataset. We also evaluate the
impact of skip connections and resolution on accuracy and efficiency.
Table~\ref{tab:ablation_table} compares the accuracy, speedup (relative to
running the teacher on every frame), fraction of frames used for supervision, and
number of FLOPS (floating point operations) for both training and inference on each
of the variants.  The baseline is \jitnet\ 0.8, the online distillation
algorithm run with an accuracy threshold of 0.8. For \jitnet\ 0.8, the maximum
updates, minimum stride, and learning rate were set to 8, 8 and 0.01
respectively. We vary one parameter at a time, and each column in the table
corresponds to a variation of the \jitnet\ 0.8 baseline.

\textbf{Learning rate:} High learning rates allow for faster adaptation.
Therefore, we chose the highest learning rate at which online training is stable
for all our experiments. As one can see in Table~\ref{tab:ablation_table}, a
lower learning rate of 0.001 reduces both accuracy and speedup. Increasing the
learning rate to 0.1 destabilizes training and yields low accuracy.

\textbf{Max updates and stride:} The number of updates needed on a single frame
depends on how much the model can learn from one frame, and how useful that
information is in the immediate future. Increasing the number of updates leads
to overfitting, reducing accuracy while increasing speedup, and reducing the
number of teacher samples used. This suggests some room for improvement in
choosing how many updates to perform on a given frame over our simple accuracy-based
heuristic. As one would expect, increasing and decreasing minimum stride
increase and decrease accuracy respectively. 

\textbf{\jitnet\ capacity:} Intuitively, as the capacity of the
student architecture is increased, the student model should require less help
from the teacher. We verify this by varying the width of \jitnet\ (the number of
channels in each layer), and observe that a smaller capacity network (width 0.5)
requires more supervision from the teacher, and also results in a significant drop
in accuracy.  Doubling \jitnet\ width improves overall accuracy and
reduces the number of teacher samples used. However, overall speedup is lower
than the baseline due to the increased inference and training cost of the wider
\jitnet\ model.

\textbf{\jitnet\ resolution:} High resolution is necessary for maintaining high accuracy on video
streams that have small objects. When the input resolution to \jitnet\ is
halved (scale 0.5), there is an overall increase in the number of frames on
which teacher supervision is used, and also a drop in accuracy relative
to the baseline. However, reducing the resolution to 75\% (scale 0.75) retains
high accuracy while being slightly faster than the baseline. This suggests the
possibility of varying resolution based on the contents in a video stream, which
could be explored in the future.

\textbf{Skip connections:} We added encoder-to-decoder skip connections to
facilitate better gradient propagation and make \jitnet\ suitable for fast
online adaptation. We evaluate the impact of the skip connections by removing
them (Table~\ref{tab:ablation_table}, No Skip). \jitnet\ without skip
connections requires more teacher samples and adaptation, reducing both
accuracy and speedup relative to the baseline.

Overall, the online distillation algorithm is reasonably robust to different
parameter settings and provides a range of options for accuracy vs. efficiency. 

\subsection{MobileNet Student}

We compare \jitnet\ with a popular efficiency-oriented
MobileNetV2\,\cite{sandler2018inverted, tfmodelzoo} architecture in the context
of online distillation. Table~\ref{tab:ablation_table} shows online accuracy and
speedup of online distillation when the MobileNetV2 architecture is used as the
student. The two MobileNetV2 variants produce outputs at $1/8^{th}$ and
$1/16^{th}$ of the input resolution. As one can see, the higher resolution
variant of the MobileNetV2 architecture is significantly slower and has lower
accuracy than the \jitnet\ baseline. Even the lower resolution (scale
0.75) version of \jitnet\ has higher accuracy and speedup compared to the
MobileNetV2 student. These results demonstrate that off-the-shelf models can be
used in our online distillation framework. However, \jitnet\ provides a better
accuracy vs. efficiency spectrum compared to MobileNetV2 for online distillation,
since it is designed for fast adaptation. Note that we measure FLOPS, which is a
platform-agnostic metric, to ensure fair comparison, since wall-clock time
(MobileNetV2 takes 38ms for inference compared to 7ms for \jitnet\ on a Nvidia
V100 GPU) depends on various factors, including target platform, underlying
libraries, and specific implementation. 

\subsection{DAVIS Evaluation}

\begin{table}[t!]\centering \small
    \begin{tabular}{L{1.75cm}C{0.65cm}C{0.65cm}C{0.85cm}C{0.65cm}C{0.65cm}C{0.65cm}} \toprule
       Method   & JM & JR & JD & FM & FR & FD \\ \hline
       JITNet A & 0.642  & 0.731    & 0.238   & 0.680  & 0.761    & 0.235   \\
       JITNet B & 0.796  & 0.927    & 0.018   & 0.798  & 0.904    & 0.060   \\
       JITNet C & 0.811  & 0.924    & -0.004  & 0.831  & 0.913    & 0.004   \\
       OSVOS-S~\cite{osvoss}  & 0.856  & 0.968    & 0.055   & 0.875  & 0.959    & 0.082   \\
       OSVOS~\cite{caelles2017one}    & 0.798  & 0.936    & 0.149   & 0.806  & 0.926    & 0.150 \\ \bottomrule 
       \end{tabular}
       \caption{Accuracy comparison of different methods using the \jitnet\
       architecture and recent methods for semi-supervised video object
       segmentation on the DAVIS 2016 benchmark.}
       \label{tab:davis}
\end{table}

Online distillation as a technique can be used to mimic an accurate teacher
model with a compact model, improving runtime efficiency. The main focus
of this work is to demonstrate the viability of the online distillation technique for semantic
segmentation on streams captured from typical use case scenarios. In this section, we show
preliminary results on the viability of online distillation combined with the \jitnet\
architecture for accelerating semi-supervised video object segmentation methods.
Specifically, we evaluate how the \jitnet\ architecture can be combined with
state-of-the-art methods such as OSVOS-S~\cite{osvoss}.

We evaluate three different configurations of \jitnet\, at varying levels
of supervision. In configuration A, we train \jitnet\ on only the first ground
truth frame of each sequence, and evaluate \jitnet\ over the rest of the frames
in the sequence without any additional supervision (the standard video object
segmentation task). On many sequences in DAVIS, object appearance changes
significantly and requires prior knowledge of the object shape. Note that \jitnet\
is a very low capacity model designed for online training, and cannot encode such
priors. Configuration A is not an online distillation scenario, but even with
its low capacity, the \jitnet\ architecture trained on just the first frame yields
reasonable results. 

Recent methods such as OSVOS-S~\cite{osvoss} leverage instance segmentation models
such as Mask R-CNN for providing priors on object shape every frame. We take a
similar approach in configuration B, where the goal is to mimic the expensive
OSVOS-S model. We train \jitnet\ on the first ground truth frame, then adapt
using segmentation predictions from OSVOS-S~\cite{osvoss} every 16 frames.  Note
that in configuration B, our combined approach does not use additional ground
truth, since OSVOS-S predictions are made using only the first ground truth
frame. Finally, in configuration C, we train on the first ground truth frame, and adapt
on the ground truth mask every 16 frames. This gives an idea of how the quality
of the teacher effects online distillation.

We use the validation set of the DAVIS 2016~\cite{DAVIS} dataset for our evaluation.
The dataset contains 50 video sequences of 3455 frames total, each labeled with
pixel-accurate segmentation masks for a single foreground object. We evaluate
using the main DAVIS metrics: region similarity J and contour accuracy F, with
precision, recall, and decay over time for both. We present metrics over the
entire DAVIS 2016 validation set for all three \jitnet\ configurations,
alongside a subset of state-of-the-art video object segmentation approaches. In
all configurations, we start with \jitnet\ pre-trained on
YouTube-VOS~\cite{xu2018youtube}, with max updates per frame set to 500,
accuracy threshold set to 0.95, and use standard data augmentation (flipping,
random noise, blurring, rotation). \jitnet\ A performs similarly to
OFL~\cite{ofl}, a flow-based approach for video object segmentation, while
\jitnet\ B, using OSVOS-S predictions, performs comparably to OSVOS, with
significantly lower runtime cost. Finally, \jitnet\ C, which uses ground truth
masks for adaptation, performs comparably to only using OSVOS-S predictions.
This suggests that even slightly noisy supervision suffices for online
distillation. Overall, these results are encouraging with regards to further work into
exploring architectures well suited for online training. 

\subsection{Offline Training Details}
\paragraph{\jitnet\ COCO pre-training:} All \jitnet\ models used in our
experiments are pre-trained on the COCO dataset. We convert the COCO instance
mask labels into semantic segmentation labels by combining all the instance
masks of each class for each image. We train the model
on all 80 classes. The model is trained on 4 GPUs with batch size 24 (6 per
GPU) using an Adam optimizer with a starting learning rate of 0.1 and a step decay
schedule (reduces learning rate to 1/10th of current rate every 10 epochs) for
30 epochs. 

\paragraph{\jitnet\ offline oracle training:} All offline oracle models are
initialized using the COCO pre-trained model and trained on the specialized
dataset for each video using the same training setup as COCO, i.e., same number
of GPUs, batch size, optimizer, and learning rate schedule. However, each of the
specialized datasets is about 6000 images, 20$\times$ smaller than the COCO
dataset.

\subsection{Standalone Semantic Segmentation}
The \jitnet\ architecture is specifically designed with low capacity so that it
can support both fast training and inference. To understand the accuracy vs.
efficiency trade-off relative to other architectures such as
MobileNetV2\,\cite{sandler2018inverted, tfmodelzoo}, we trained a \jitnet\ model
with twice the number of channels and encoder/decoder blocks
than the one used in the paper. This modified architecture is 1.5$\times$ faster
than the semantic segmentation architecture based on MobileNetV2. The larger
\jitnet\ gives a mean IoU of 67.34 on the
cityscapes\,\cite{cordts2016cityscapes} validation set and compares favorably
with the 70.71 mean IoU of the MobileNetV2 based model\,\cite{tfmodelzoo}. We
started with the larger \jitnet\ architecture in the online distillation
experiments, but lowered the capacity even further, with half the number of channels
and encoder/decoder blocks, since it provided a better cost vs. accuracy
trade-off for online distillation.

\renewcommand{\arraystretch}{0.8}
\begin{table*}[t!]
    \small \centering
    \begin{tabular}{L{2.8cm}C{1.25cm}C{1.75cm}C{1.5cm}C{2.5cm}C{2.5cm}C{2.5cm}}\toprule
         &  Offline & \multicolumn{2}{c}{Flow} &
        \multicolumn{3}{c}{Online Distillation} \\
        \cmidrule{3-4} \cmidrule{5-7}
        Video &  Oracle & {Slow (2.2$\times$)} & {Fast (3.2$\times$)} & {\jitnet\ 0.7} &
        {\jitnet\ 0.8} & {\jitnet\ 0.9} \\
        & (20\%)       & (12.5\%) & ( 6.2\%) & & & \\
\arrayrulecolor{black!30}\midrule
{\textbf{Overall}}& 80.3& 76.6& 65.2& 75.5 (17.4$\times$, 3.2\%)& 78.6 (13.5$\times$, 4.7\%)& 82.5 ($\times$7.5, 8.4\%)\\
\arrayrulecolor{black!30}\midrule
         \multicolumn{7}{c}{Category Averages}\\
\arrayrulecolor{black!30}\midrule
{Sports (Fixed)}& 87.5& 81.2& 71.0& 80.8(36.7$\times$, 1.6\%)& 82.8(33.3$\times$, 1.8\%)& 87.6(16.1$\times$, 5.1\%)\\
\arrayrulecolor{black!30}\midrule
{Sports (Moving)}& 82.2& 72.6& 59.8& 76.0(31.4$\times$, 2.1\%)& 79.3(22.2$\times$, 3.6\%)& 84.1(9.3$\times$, 9.1\%)\\
\arrayrulecolor{black!30}\midrule
{Sports (Ego)}& 72.3& 69.4& 55.1& 65.0(21.1$\times$, 3.7\%)& 70.2(14.1$\times$, 6.0\%)& 75.0(7.7$\times$, 10.4\%)\\
\arrayrulecolor{black!30}\midrule
{Animals}& 89.0& 83.2& 73.4& 82.9(33.1$\times$, 1.9\%)& 84.3(30.1$\times$, 2.2\%)& 87.6(22.0$\times$, 4.4\%)\\
\arrayrulecolor{black!30}\midrule
{Traffic}& 82.3& 82.6& 74.0& 79.1(18.4$\times$, 4.6\%)& 82.1(13.3$\times$, 7.1\%)& 84.3(8.4$\times$, 10.1\%)\\
\arrayrulecolor{black!30}\midrule
{Driving/Walking}& 50.6& 69.3& 55.9& 59.6(9.0$\times$, 8.6\%)& 63.9(7.6$\times$, 10.5\%)& 66.6(6.7$\times$, 11.9\%)\\
\arrayrulecolor{black!30}\midrule
         \multicolumn{7}{c}{Individual Video Streams}\\
\arrayrulecolor{black!30}\midrule
{Badminton (P)}& 83.1& 83.2& 72.9& 77.1($36.7\times$, 1.6\%)& 80.0($32.6\times$, 1.8\%)& 87.3(9.8$\times$, 7.9\%)\\
\arrayrulecolor{black!30}\midrule
{Squash (P)}& 88.4& 70.0& 56.5& 80.9($37.0\times$, 1.6\%)& 82.5($35.7\times$, 1.7\%)& 86.0(21.3$\times$, 3.2\%)\\
\arrayrulecolor{black!30}\midrule
{Table Tennis (P)}& 89.4& 84.8& 75.4& 81.5($37.2\times$, 1.6\%)& 83.5($36.7\times$, 1.6\%)& 88.3(20.3$\times$, 3.4\%)\\
\arrayrulecolor{black!30}\midrule
{Softball (P)}& 89.2& 86.7& 79.2& 83.8($36.0\times$, 1.7\%)& 85.3($28.2\times$, 2.3\%)& 88.8(13.1$\times$, 5.7\%)\\
\arrayrulecolor{black!30}\midrule
{Hockey (P)}& 81.9& 68.0& 54.5& 75.7($31.5\times$, 2.0\%)& 79.0($18.5\times$, 3.8\%)& 84.2(7.3$\times$, 10.8\%)\\
\arrayrulecolor{black!30}\midrule
{Soccer (P)}& 80.0& 68.3& 54.6& 75.2($33.2\times$, 1.8\%)& 79.0($18.9\times$, 3.7\%)& 83.7(7.3$\times$, 10.8\%)\\
\arrayrulecolor{black!30}\midrule
{Tennis (P)}& 87.3& 80.1& 67.5& 81.1($35.9\times$, 1.6\%)& 82.5($32.2\times$, 1.9\%)& 87.2(15.4$\times$, 4.8\%)\\
\arrayrulecolor{black!30}\midrule
{Volleyball (P)}& 82.3& 82.9& 73.0& 76.4($34.3\times$, 1.7\%)& 80.3($21.1\times$, 3.2\%)& 85.0(8.4$\times$, 9.2\%)\\
\arrayrulecolor{black!30}\midrule
{Ice Hockey (P)}& 79.0& 72.8& 60.2& 72.0($30.8\times$, 2.0\%)& 76.3($19.1\times$, 3.7\%)& 81.8(7.3$\times$, 10.7\%)\\
\arrayrulecolor{black!30}\midrule
{Kabaddi (P)}& 88.2& 78.9& 66.7& 83.8($37.2\times$, 1.6\%)& 84.5($35.6\times$, 1.7\%)& 87.9(12.1$\times$, 6.3\%)\\
\arrayrulecolor{black!30}\midrule
{Figure Skating (P)}& 84.3& 54.8& 37.9& 72.3($24.3\times$, 2.8\%)& 76.0($17.6\times$, 4.1\%)& 83.5(8.3$\times$, 9.4\%)\\
\arrayrulecolor{black!30}\midrule
{Drone (P)}& 74.5& 70.5& 58.5& 70.8($23.7\times$, 2.8\%)& 76.6($10.7\times$, 7.2\%)& 79.9(6.3$\times$, 12.5\%)\\
\arrayrulecolor{black!30}\midrule
{Elephant (E)}& 93.3& 91.0& 85.3& 92.7($37.1\times$, 1.6\%)& 92.8($37.2\times$, 1.6\%)& 93.6(36.6$\times$, 1.6\%)\\
\arrayrulecolor{black!30}\midrule
{Birds (B)}& 92.0& 80.0& 68.0& 85.3($37.0\times$, 1.6\%)& 85.7($36.8\times$, 1.6\%)& 87.9(33.7$\times$, 1.8\%)\\
\arrayrulecolor{black!30}\midrule
{Giraffe (P,G)}& 85.5& 79.6& 69.2& 82.8($32.1\times$, 1.9\%)& 84.1($26.4\times$, 2.5\%)& 87.6(11.4$\times$, 6.6\%)\\
\arrayrulecolor{black!30}\midrule
{Dog (P,D,C)}& 86.1& 80.4& 71.1& 78.4($29.3\times$, 2.2\%)& 81.2($21.4\times$, 3.2\%)& 86.5(9.2$\times$, 8.4\%)\\
\arrayrulecolor{black!30}\midrule
{Horse (P,H)}& 87.9& 84.9& 73.4& 75.3($30.1\times$, 2.1\%)& 77.7($28.6\times$, 2.2\%)& 82.7(19.2$\times$, 3.6\%)\\
\arrayrulecolor{black!30}\midrule
{Ego Ice Hockey (P)}& 68.8& 56.7& 39.6& 56.3($31.1\times$, 2.0\%)& 59.3($20.1\times$, 3.4\%)& 67.0(7.8$\times$, 10.0\%)\\
\arrayrulecolor{black!30}\midrule
{Ego Basketball (P,C)}& 68.4& 70.5& 56.2& 59.8($13.1\times$, 5.7\%)& 67.9($9.9\times$, 7.8\%)& 70.1(7.4$\times$, 10.7\%)\\
\arrayrulecolor{black!30}\midrule
{Ego Dodgeball (P)}& 82.1& 75.5& 60.4& 74.3($26.6\times$, 2.5\%)& 79.5($20.3\times$, 3.4\%)& 84.2(9.5$\times$, 8.2\%)\\
\arrayrulecolor{black!30}\midrule
{Ego Soccer (P)}& 71.3& 72.9& 58.2& 66.3($14.8\times$, 5.0\%)& 72.1($9.5\times$, 8.1\%)& 78.3(7.2$\times$, 10.9\%)\\
\arrayrulecolor{black!30}\midrule
{Biking (P,B)}& 70.7& 71.6& 61.3& 68.2($19.8\times$, 3.5\%)& 72.3($10.4\times$, 7.3\%)& 75.3(6.4$\times$, 12.4\%)\\
\arrayrulecolor{black!30}\midrule
{Streetcam1 (P,C)}& 86.0& 76.8& 65.3& 79.1($25.2\times$, 2.5\%)& 82.1($19.1\times$, 3.6\%)& 85.5(13.8$\times$, 5.2\%)\\
\arrayrulecolor{black!30}\midrule
{Streetcam2 (P,C)}& 82.2& 82.1& 72.9& 76.1($15.9\times$, 4.6\%)& 79.7($10.1\times$, 7.6\%)& 83.7(6.5$\times$, 12.2\%)\\
\arrayrulecolor{black!30}\midrule
{Jackson Hole (P,C)}& 76.5& 77.9& 67.9& 75.7($12.8\times$, 5.9\%)& 78.0($9.4\times$, 8.3\%)& 79.2(7.4$\times$, 10.7\%)\\
\arrayrulecolor{black!30}\midrule
{Murphys (P,C,B)}& 91.9& 94.1& 91.2& 88.0($32.1\times$, 1.9\%)& 89.8($26.0\times$, 2.5\%)& 92.9(9.9$\times$, 7.8\%)\\
\arrayrulecolor{black!30}\midrule
{Samui Street (P,C,B)}& 80.6& 83.8& 76.5& 78.8($13.6\times$, 5.5\%)& 82.6($8.2\times$, 9.5\%)& 83.7(6.5$\times$, 12.2\%)\\
\arrayrulecolor{black!30}\midrule
{Toomer (P,C)}& 76.6& 81.1& 70.4& 76.9($10.7\times$, 7.2\%)& 80.3($7.0\times$, 11.3\%)& 80.5(6.4$\times$, 12.4\%)\\
\arrayrulecolor{black!30}\midrule
{Driving (P,C,B)}& 51.1& 72.2& 59.7& 63.8($8.9\times$, 8.8\%)& 68.2($6.9\times$, 11.5\%)& 66.7(6.4$\times$, 12.4\%)\\
\arrayrulecolor{black!30}\midrule
{Walking (P,C,B)}& 50.2& 66.4& 52.1& 55.4($9.1\times$, 8.5\%)& 59.6($8.2\times$, 9.5\%)& 66.4(7.0$\times$, 11.3\%)\\
\arrayrulecolor{black}\bottomrule
    \end{tabular}
    \vspace{-0.5em}
    \caption{Comparison of accuracy (mean IoU over all the classes excluding
    background), runtime speedup relative to MRCNN (where applicable), and the
    fraction of frames where MRCNN is run.  Classes
    present in each video are denoted by letters (A - Auto, Bi - Bird, Bk - Bike, D -
    Dog, E - Elephant, G - Giraffe, H - Horse, P - Person).  Overall, online distillation using
    \jitnet\ provides a better accuracy/efficiency tradeoff than baseline methods.}
    \label{tab:main_table}
\end{table*}

\subsection{Additional Results}
 Table~\ref{tab:main_table} gives the accuracy and performance of online
 distillation for each individual video stream we used in our evaluation, using
 \jitnet\ at three different accuracy thresholds: \jitnet~0.7, 0.8, and 0.9.

%% file: paper.bbl
\begin{thebibliography}{10}\itemsep=-1pt

\bibitem{badrinarayanan2017segnet}
Vijay Badrinarayanan, Alex Kendall, and Roberto Cipolla.
\newblock {SegNet}: A deep convolutional encoder-decoder architecture for image
  segmentation.
\newblock {\em IEEE Transactions on Pattern Analysis and Machine Intelligence},
  39(12):2481--2495, 2017.

\bibitem{bucilua2006model}
Cristian Buciluǎ, Rich Caruana, and Alexandru Niculescu-Mizil.
\newblock Model compression.
\newblock In {\em Proceedings of the 12th ACM SIGKDD International Conference
  on Knowledge Discovery and Data Mining}, pages 535--541. ACM, 2006.

\bibitem{inplaceabn}
Samuel Bulo, Lorenzo Porzi, and Peter Kontschieder.
\newblock In-place activated batchnorm for memory-optimized training of {DNN}s.
\newblock In {\em Proceedings of the IEEE Conference on Computer Vision and
  Pattern Recognition}, 2018.

\bibitem{caelles2017one}
Sergi Caelles, Kevis-Kokitsi Maninis, Jordi Pont-Tuset, Laura Leal-Taix{\'e},
  Daniel Cremers, and Luc Van~Gool.
\newblock One-shot video object segmentation.
\newblock In {\em Proceedings of the IEEE Conference on Computer Vision and
  Pattern Recognition (CVPR)}. IEEE, 2017.

\bibitem{cesa2006prediction}
Nicolo Cesa-Bianchi and G{\'a}bor Lugosi.
\newblock {\em Prediction, Learning, and Games}.
\newblock Cambridge University Press, 2006.

\bibitem{chen2017rethinking}
Liang-Chieh Chen, George Papandreou, Florian Schroff, and Hartwig Adam.
\newblock Rethinking atrous convolution for semantic image segmentation.
\newblock {\em arXiv preprint arXiv:1706.05587}, 2017.

\bibitem{cordts2016cityscapes}
Marius Cordts, Mohamed Omran, Sebastian Ramos, Timo Rehfeld, Markus Enzweiler,
  Rodrigo Benenson, Uwe Franke, Stefan Roth, and Bernt Schiele.
\newblock The cityscapes dataset for semantic urban scene understanding.
\newblock In {\em Proceedings of the IEEE conference on computer vision and
  pattern recognition}, pages 3213--3223, 2016.

\bibitem{detectron}
{FAIR}.
\newblock {Detectron Mask R-CNN}.
\newblock \url{https://github.com/facebookresearch/Detectron}, 2018.

\bibitem{pmlr-v70-finn17a}
Chelsea Finn, Pieter Abbeel, and Sergey Levine.
\newblock Model-agnostic meta-learning for fast adaptation of deep networks.
\newblock In Doina Precup and Yee~Whye Teh, editors, {\em Proceedings of the
  34th International Conference on Machine Learning}, volume~70 of {\em
  Proceedings of Machine Learning Research}, pages 1126--1135, International
  Convention Centre, Sydney, Australia, 06--11 Aug 2017. PMLR.

\bibitem{fleuret2008multicamera}
Francois Fleuret, Jerome Berclaz, Richard Lengagne, and Pascal Fua.
\newblock Multicamera people tracking with a probabilistic occupancy map.
\newblock {\em IEEE Transactions on Pattern Analysis and Machine Intelligence},
  30(2):267--282, 2008.

\bibitem{gadde2017semantic}
Raghudeep Gadde, Varun Jampani, and Peter~V Gehler.
\newblock Semantic video {CNNs} through representation warping.
\newblock {\em CoRR, abs/1708.03088}, 2017.

\bibitem{goodman1988stability}
Jonathan Goodman, Albert~G Greenberg, Neal Madras, and Peter March.
\newblock Stability of binary exponential backoff.
\newblock {\em Journal of the ACM (JACM)}, 35(3):579--602, 1988.

\bibitem{hare2016struck}
Sam Hare, Stuart Golodetz, Amir Saffari, Vibhav Vineet, Ming-Ming Cheng,
  Stephen~L Hicks, and Philip~HS Torr.
\newblock Struck: Structured output tracking with kernels.
\newblock {\em IEEE Transactions on Pattern Analysis and Machine Intelligence},
  38(10):2096--2109, 2016.

\bibitem{he2017mask}
Kaiming He, Georgia Gkioxari, Piotr Doll{\'a}r, and Ross Girshick.
\newblock Mask {R-CNN}.
\newblock In {\em Proceedings of the IEEE International Conference on Computer
  Vision (ICCV)}, pages 2980--2988. IEEE, 2017.

\bibitem{resnet}
Kaiming He, Xiangyu Zhang, Shaoqing Ren, and Jian Sun.
\newblock Deep residual learning for image recognition.
\newblock In {\em Proceedings of the IEEE Conference on Computer Vision and
  Pattern Recognition (CVPR)}, June 2016.

\bibitem{henriques2015high}
Jo{\~a}o~F Henriques, Rui Caseiro, Pedro Martins, and Jorge Batista.
\newblock High-speed tracking with kernelized correlation filters.
\newblock {\em IEEE Transactions on Pattern Analysis and Machine Intelligence},
  37(3):583--596, 2015.

\bibitem{hinton2015distilling}
Geoffrey Hinton, Oriol Vinyals, and Jeff Dean.
\newblock Distilling the knowledge in a neural network.
\newblock {\em arXiv preprint arXiv:1503.02531}, 2015.

\bibitem{hong2015online}
Seunghoon Hong, Tackgeun You, Suha Kwak, and Bohyung Han.
\newblock Online tracking by learning discriminative saliency maps with
  convolutional neural network.
\newblock In {\em International Conference on Machine Learning}, pages
  597--606, 2015.

\bibitem{ilg2017flownet}
Eddy Ilg, Nikolaus Mayer, Tonmoy Saikia, Margret Keuper, Alexey Dosovitskiy,
  and Thomas Brox.
\newblock {FlowNet 2.0}: Evolution of optical flow estimation with deep
  networks.
\newblock In {\em Proceedings of the IEEE Conference on Computer Vision and
  Pattern Recognition (CVPR)}, pages 2462--2470, 2017.

\bibitem{Jiang:2018:chameleon}
Junchen Jiang, Ganesh Ananthanarayanan, Peter Bodik, Siddhartha Sen, and Ion
  Stoica.
\newblock Chameleon: Scalable adaptation of video analytics.
\newblock In {\em Proceedings of the 2018 Conference of the ACM Special
  Interest Group on Data Communication}, SIGCOMM '18, pages 253--266, New York,
  NY, USA, 2018. ACM.

\bibitem{kalal2012tracking}
Zdenek Kalal, Krystian Mikolajczyk, and Jiri Matas.
\newblock Tracking-{L}earning-{D}etection.
\newblock {\em IEEE Transactions on Pattern Analysis and Machine Intelligence},
  34(7):1409--1422, 2012.

\bibitem{kang2017noscope}
Daniel Kang, John Emmons, Firas Abuzaid, Peter Bailis, and Matei Zaharia.
\newblock Noscope: Optimizing neural network queries over video at scale.
\newblock {\em Proceedings of the VLDB Endowment}, 10(11):1586--1597, 2017.

\bibitem{koutnik2014clockwork}
Jan Koutnik, Klaus Greff, Faustino Gomez, and Juergen Schmidhuber.
\newblock A clockwork {RNN}.
\newblock In {\em Proceedings of the International Conference on Machine
  Learning}, pages 1863--1871, 2014.

\bibitem{kuehne2011hmdb}
Hildegard Kuehne, Hueihan Jhuang, Est{\'\i}baliz Garrote, Tomaso Poggio, and
  Thomas Serre.
\newblock {HMDB}: A large video database for human motion recognition.
\newblock In {\em Proceedings of the IEEE International Conference on Computer
  Vision}, pages 2556--2563. IEEE, 2011.

\bibitem{lin2014microsoft}
Tsung-Yi Lin, Michael Maire, Serge Belongie, James Hays, Pietro Perona, Deva
  Ramanan, Piotr Doll{\'a}r, and C~Lawrence Zitnick.
\newblock {Microsoft COCO}: Common objects in context.
\newblock In {\em European Conference on Computer Vision}, pages 740--755.
  Springer, 2014.

\bibitem{liu2013intelligent}
Honghai Liu, Shengyong Chen, and Naoyuki Kubota.
\newblock Intelligent video systems and analytics: A survey.
\newblock {\em IEEE Trans. Industrial Informatics}, 9(3):1222--1233, 2013.

\bibitem{lu2013learning}
Wei-Lwun Lu, Jo-Anne Ting, James~J Little, and Kevin~P Murphy.
\newblock Learning to track and identify players from broadcast sports videos.
\newblock {\em IEEE transactions on pattern analysis and machine intelligence},
  35(7):1704--1716, 2013.

\bibitem{ma2015hierarchical}
Chao Ma, Jia-Bin Huang, Xiaokang Yang, and Ming-Hsuan Yang.
\newblock Hierarchical convolutional features for visual tracking.
\newblock In {\em Proceedings of the IEEE International Conference on Computer
  Vision (ICCV)}, pages 3074--3082, 2015.

\bibitem{osvoss}
K. Maninis, S. Caelles, Y. Chen, J. Pont-{T}uset, L. Leal-{T}aixe, D. Cremers,
  and L. Van~Gool.
\newblock Video object segmentation without temporal information.
\newblock 2018.

\bibitem{mnih2016asynchronous}
Volodymyr Mnih, Adria~Puigdomenech Badia, Mehdi Mirza, Alex Graves, Timothy
  Lillicrap, Tim Harley, David Silver, and Koray Kavukcuoglu.
\newblock Asynchronous methods for deep reinforcement learning.
\newblock In {\em International conference on machine learning}, pages
  1928--1937, 2016.

\bibitem{mnih2015human}
Volodymyr Mnih, Koray Kavukcuoglu, David Silver, Andrei~A Rusu, Joel Veness,
  Marc~G Bellemare, Alex Graves, Martin Riedmiller, Andreas~K Fidjeland, Georg
  Ostrovski, et~al.
\newblock Human-level control through deep reinforcement learning.
\newblock {\em Nature}, 518(7540):529, 2015.

\bibitem{nam2016learning}
Hyeonseob Nam and Bohyung Han.
\newblock Learning multi-domain convolutional neural networks for visual
  tracking.
\newblock In {\em Proceedings of the IEEE Conference on Computer Vision and
  Pattern Recognition (CVPR)}, pages 4293--4302. IEEE, 2016.

\bibitem{perazzi2017learning}
Federico Perazzi, Anna Khoreva, Rodrigo Benenson, Bernt Schiele, and Alexander
  Sorkine-Hornung.
\newblock Learning video object segmentation from static images.
\newblock In {\em Proceedings of the IEEE Conference on Computer Vision and
  Pattern Recognition (CVPR)}, 2017.

\bibitem{DAVIS}
F. Perazzi, J. Pont-{T}uset, B. Mc{W}illiams, L. Van~{G}ool, M. Gross, and A.
  Sorkine-{H}ornung.
\newblock A benchmark dataset and evaluation methodology for video object
  segmentation.
\newblock In {\em Proceedings of the IEEE Conference on Computer Vision and
  Pattern Recognition (CVPR)}, 2016.

\bibitem{ilsvrc15}
Olga Russakovsky, Jia Deng, Hao Su, Jonathan Krause, Sanjeev Satheesh, Sean Ma,
  Zhiheng Huang, Andrej Karpathy, Aditya Khosla, Michael Bernstein,
  Alexander~C. Berg, and Li Fei-Fei.
\newblock {ImageNet Large Scale Visual Recognition Challenge}.
\newblock {\em International Journal of Computer Vision (IJCV)},
  115(3):211--252, 2015.

\bibitem{sandler2018inverted}
Mark Sandler, Andrew Howard, Menglong Zhu, Andrey Zhmoginov, and Liang-Chieh
  Chen.
\newblock {MobileNetV2}: Inverted residuals and linear bottlenecks.
\newblock In {\em Proceedings of the IEEE Conference on Computer Vision and
  Pattern Recognition (CVPR)}, 2018.

\bibitem{schulman2017proximal}
John Schulman, Filip Wolski, Prafulla Dhariwal, Alec Radford, and Oleg Klimov.
\newblock Proximal policy optimization algorithms.
\newblock {\em arXiv preprint arXiv:1707.06347}, 2017.

\bibitem{shalev2012online}
Shai Shalev-Shwartz et~al.
\newblock Online learning and online convex optimization.
\newblock {\em Foundations and Trends{\textregistered} in Machine Learning},
  4(2):107--194, 2012.

\bibitem{shelhamer2016clockwork}
Evan Shelhamer, Kate Rakelly, Judy Hoffman, and Trevor Darrell.
\newblock Clockwork convnets for video semantic segmentation.
\newblock In {\em Proceedings of the European Conference on Computer Vision
  (ECCV)}, pages 852--868. Springer, 2016.

\bibitem{Shen:2017:adaptivedetection}
H. Shen, S. Han, M. Philipose, and A. Krishnamurthy.
\newblock Fast video classification via adaptive cascading of deep models.
\newblock In {\em Proceedings of the IEEE Conference on Computer Vision and
  Pattern Recognition (CVPR)}, July 2017.

\bibitem{Shocher:2018:superres}
Assaf Shocher, Nadav Cohen, and Michal Irani.
\newblock “{Z}ero-shot” super-resolution using deep internal learning.
\newblock In {\em Proceedings of the IEEE Conference on Computer Vision and
  Pattern Recognition}, pages 3118--3126, 2018.

\bibitem{soomro2012ucf101}
Khurram Soomro, Amir~Roshan Zamir, and Mubarak Shah.
\newblock {UCF101}: A dataset of 101 human actions classes from videos in the
  wild.
\newblock {\em arXiv preprint arXiv:1212.0402}, 2012.

\bibitem{sun2018pwcnet}
Deqing Sun, Xiaodong Yang, Ming-{Y}u Liu, and Jan Kautz.
\newblock {PWC-Net: CNN}s for optical flow using pyramid, warping, and cost
  volume.
\newblock In {\em Proceedings of the IEEE Conference on Computer Vision and
  Pattern Recognition (CVPR)}, 2018.

\bibitem{tfmodelzoo}
Tensor{F}low.
\newblock {TensorFlow DeepLab Model Zoo}.
\newblock
  \url{https://github.com/tensorflow/models/blob/master/research/deeplab/g3doc/model_zoo.md},
  2018.

\bibitem{ofl}
Yi-Hsuan Tsai, Ming-Hsuan Yang, and Michael Black.
\newblock Video segmentation via object flow.
\newblock In {\em Proceedings of the IEEE Conference on Computer Vision and
  Pattern Recognition}, 2016.

\bibitem{voigtlaender:2017:onAVOS}
Paul Voigtlaender and Bastian Leibe.
\newblock Online adaptation of convolutional neural networks for video object
  segmentation.
\newblock In {\em BMVC}, 2017.

\bibitem{wang2015visual}
Lijun Wang, Wanli Ouyang, Xiaogang Wang, and Huchuan Lu.
\newblock Visual tracking with fully convolutional networks.
\newblock In {\em Proceedings of the IEEE International Conference on Computer
  Vision (ICCV)}, pages 3119--3127, 2015.

\bibitem{xu2018youtube}
Ning Xu, Linjie Yang, Yuchen Fan, Dingcheng Yue, Yuchen Liang, Jianchao Yang,
  and Thomas Huang.
\newblock {YouTube-VOS}: A large-scale video object segmentation benchmark.
\newblock {\em arXiv preprint arXiv:1809.03327}, 2018.

\bibitem{yang2018efficient}
Linjie Yang, Yanran Wang, Xuehan Xiong, Jianchao Yang, and Aggelos~K
  Katsaggelos.
\newblock Efficient video object segmentation via network modulation.
\newblock {\em Proceedings of the International Conference on Robotics and
  Automation}, 2018.

\bibitem{zhu2017deep}
Xizhou Zhu, Yuwen Xiong, Jifeng Dai, Lu Yuan, and Yichen Wei.
\newblock Deep feature flow for video recognition.
\newblock In {\em Proceedings of the IEEE Conference on Computer Vision and
  Pattern Recognition (CVPR)}, volume~2, page~7, 2017.

\end{thebibliography}
